\documentclass{article}
\usepackage[numbers,sort&compress]{natbib}
\usepackage[final]{report}
\usepackage[utf8]{inputenc}
\usepackage[T1]{fontenc}
\usepackage{amsmath,amsfonts,bm,amsthm}
\usepackage{graphicx,wrapfig,float}
\usepackage{needspace}
\usepackage{booktabs,colortbl}
\usepackage{microtype}
\usepackage{xcolor}
\usepackage{hyperref,url}
\usepackage{etoc}
\usepackage[most]{tcolorbox}

\DeclareMathOperator*{\argmin}{arg\,min}

\newcommand{\grouprule}[1]{\cmidrule(lr){#1}}
\DeclareRobustCommand{\correspondingicon}{\tikz[x=1em,y=1em,baseline=0pt]{\draw[line width=0.055em,line join=round] (0,0) rectangle (0.9,0.6);
    \draw[line width=0.055em,line join=round] (0,0.6)--(0.45,0.27)--(0.9,0.6);}}
\definecolor{customgray}{RGB}{240,240,240}
\definecolor{citecolor}{HTML}{0071BC}
\definecolor{linkcolor}{HTML}{ED1C24}
\definecolor{takeawaybg}{HTML}{F7F9FC}
\definecolor{takeawayrule}{HTML}{5F7FA3}
\definecolor{takeawayaccent}{HTML}{315F86}

\newcounter{takeaway}
\renewcommand{\thetakeaway}{\arabic{takeaway}}
\newcommand{\takeawaytitle}[1]{Takeaway~\thetakeaway\if\relax\detokenize{#1}\relax\else: #1\fi}
\tcbset{takeawaystyle/.style={
  enhanced,breakable,colback=takeawaybg,colframe=takeawayrule,
  boxrule=0.55pt,arc=2pt,boxsep=0pt,left=6pt,right=6pt,top=5pt,bottom=5pt,
  before skip=0.50\baselineskip,after skip=0.35\baselineskip,
  borderline west={1.6pt}{0pt}{takeawayaccent}
}}
\newcommand{\takeaway}[2][]{\refstepcounter{takeaway}\begin{tcolorbox}[takeawaystyle]
  \textbf{\textcolor{takeawayaccent}{\takeawaytitle{#1}.}} #2
  \end{tcolorbox}}

\hypersetup{
  colorlinks=true,linkcolor=linkcolor,citecolor=citecolor,urlcolor=citecolor,
  pdftitle={How Far Are We from Removing the Visual Encoder? Scaling Laws for Encoder-Free Multimodal Pretraining},
  pdfauthor={Lin Chen; Bolin Ni; Qi Yang; Lan Jiang; Kun Ding; Xiaoran Fan; Hower Yang; Ying Wang; Shiming Xiang}
}

\title{How Far Are We from Removing the Visual Encoder?\\
Scaling Laws for Encoder-Free Multimodal Pretraining}
\author{\textbf{Lin Chen}\textsuperscript{1,2,3}\thanks{Work done during an internship at Tencent.}\quad
  \textbf{Bolin Ni}\textsuperscript{3}\thanks{Project lead\quad\textsuperscript{\correspondingicon}\,Corresponding author}\quad
  \textbf{Qi Yang}\textsuperscript{3}\quad
  \textbf{Lan Jiang}\textsuperscript{3}\quad
  \textbf{Kun Ding}\textsuperscript{1}\\[3pt]
  \textbf{Xiaoran Fan}\textsuperscript{3}\quad
  \textbf{Hower Yang}\textsuperscript{3}\quad
  \textbf{Ying Wang}\textsuperscript{1 \correspondingicon}\quad
  \textbf{Shiming Xiang}\textsuperscript{1,2}\\[7pt]
  {\small\textsuperscript{1}CASIA\quad
  \textsuperscript{2}UCAS\quad
  \textsuperscript{3}Foundation Model Department, Tencent}}

\begin{document}

\setcounter{footnote}{0}\maketitle
\setcounter{footnote}{0}
\etocdepthtag.toc{mainmatter}

\begin{abstract}
Most modern multimodal large language models (MLLMs) build on a pretrained visual encoder that provides a strong visual prior. Encoder-free MLLMs instead learn visual representations directly from raw pixels, offering a simple and unified architecture, but their scaling behavior has not been systematically characterized. To fill this gap, we compare scaling laws for encoder-free and encoder-based MLLMs and report three main findings:
\textbf{(1)} Removing the visual encoder shifts the compute-optimal allocation for the multimodal objective toward larger models, while leaving that for text nearly unchanged.
\textbf{(2)} The two architectures exhibit nearly overlapping loss--compute frontiers on the text objective, but diverge on the multimodal objective: encoder-free models underperform at small scales yet are predicted to catch up at around $10^{22}$ FLOPs, well within practical pretraining budgets.
\textbf{(3)} Without a visual encoder, the language model learns to take over its role via vision-specific adaptation: bidirectional interactions among visual tokens become increasingly beneficial as training compute grows, visual processing shifts toward earlier layers, and expert routing for visual tokens becomes more concentrated.
Overall, our results indicate that the advantage of the visual prior provided by a pretrained encoder diminishes with scale, positioning encoder-free architectures as a promising direction for multimodal pretraining.
\end{abstract}

\section{Introduction}

Most modern multimodal large language models (MLLMs)~\citep{liu2023llava, bai2025qwen3, team2025gemma} adopt an encoder-based architecture: a pretrained visual encoder~\citep{radford2021learning,tschannen2025siglip2} supplies the language model with semantically rich visual representations, providing a strong visual prior learned from large-scale image--text data.
To achieve a simple and unified architecture, encoder-free MLLMs remove the visual encoder and feed projected image patches directly into the decoder, which must then learn visual representations from raw pixels~\citep{fuyu-8b,diao2024unveiling,chen2024solo,lei2025scalability,diao2026pixels,team2026gemma,tuna2,thinkingmachines2026inkling}.
Although these studies show initial feasibility, the scaling behavior of encoder-free MLLMs has not been systematically characterized.

To this end, we conduct a controlled scaling study of encoder-free and encoder-based MLLMs, in which the two model families share the same sparse decoder ladder, data mixture, optimization setup, and visual-token granularity.
We fit scaling laws separately for the text and multimodal objectives to quantify how the efficiency gap between the two architectures evolves with scale.
To understand the mechanisms behind these trends, we further probe the decoder's internals and examine how it compensates for the missing visual encoder.
Our main findings are as follows.

\begin{figure}
  \centering
  \includegraphics[width=0.90\textwidth]{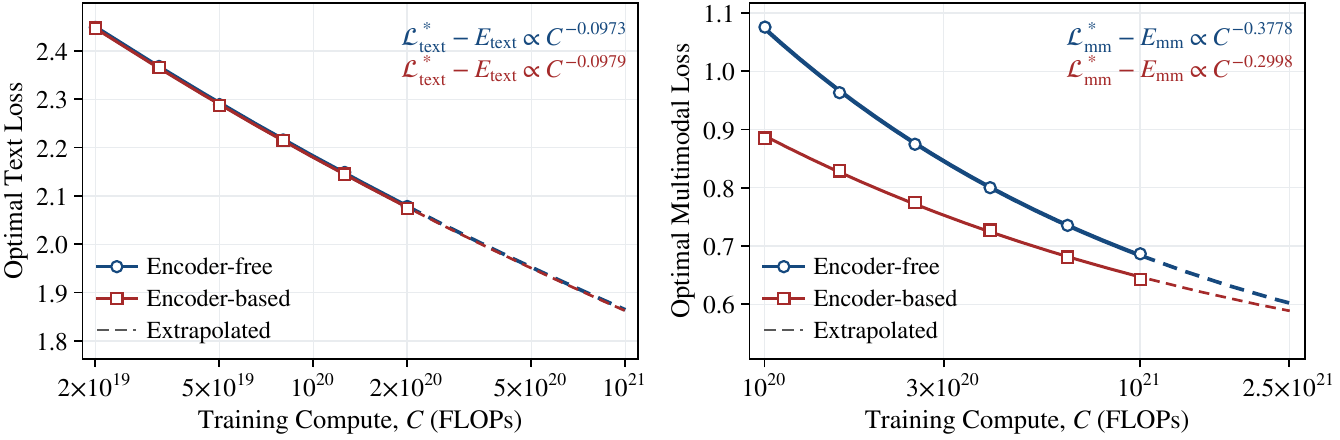}
  \vspace{-5pt}
  \caption{Compute-optimal loss frontiers for encoder-free and encoder-based models.
  \textbf{Left:} The text loss frontiers of the two architectures nearly overlap.
  \textbf{Right:} Encoder-free models have higher multimodal loss over the measured range, but their loss decreases faster with compute.}
  \vspace{-3pt}
  \label{fig:teaser}
\end{figure}

\noindent\textbf{(1) Compute-optimal encoder-free training favors larger models}~(\S\ref{sec:results:allocation}).
For the text objective, the two architectures exhibit nearly identical compute-optimal allocation trends.
In contrast, the multimodal objective shows a different pattern: removing the visual encoder increases the model allocation exponent from $a=0.464$ to $a=0.570$, shifting the optimum toward larger model scale.
This shift suggests that encoder-free models require greater decoder capacity to jointly support visual representation learning and language modeling.

\noindent\textbf{(2) Encoder-free models are predicted to catch up within practical pretraining budgets}~(\S\ref{sec:results:equal-loss}).
As shown in Fig.~\ref{fig:teaser} (Left), the two 
architectures exhibit nearly overlapping loss--compute frontiers for the text objective. In contrast, 
Fig.~\ref{fig:teaser} (Right) shows that encoder-free models require more training compute than encoder-based models to reach the same 
validation loss on the multimodal objective.
Nevertheless, their loss decreases more rapidly with compute, narrowing the gap at scale.
Extrapolating the fitted scaling laws beyond our measured range predicts that the multimodal crossover occurs on the order of $10^{22}$ FLOPs under compute-optimal allocation, and at a higher compute budget under $5\times$ overtraining.
For reference, the pretraining compute of recent flagship models, such as Kimi K2.5~\citep{team2026kimi25}, is approximately $10^{25}$ FLOPs.\footnote{Estimated using $C\approx6N_{\mathrm{active}}D$.}
Furthermore, fits on individual multimodal topics show that the crossover varies by topic, arriving earlier on topics that rely mainly on language and much later on perception-intensive ones.

\noindent\textbf{(3) The decoder takes over visual encoding via vision-specific adaptation}~(\S\ref{sec:decoder-consequences}).
The decoder increasingly relies on bidirectional attention among visual tokens, recovering the patch-level contextualization that a visual encoder would otherwise provide.
Moreover, visual token representations diverge from their inputs much earlier than in encoder-based models, so the shallow decoder layers effectively serve as an implicit visual encoding stage, whereas text tokens are processed almost identically in both architectures.
This vision-specific adaptation further extends to the MoE experts, where the routing of visual tokens becomes more concentrated, consistent with some experts taking over the vision-specific role of the visual encoder.
These adaptations suggest that encoder-free models may benefit from decoder architectures designed explicitly for native visual representation learning, rather than directly inheriting designs built for language.

Overall, encoder-free models require more training compute within the fitted range, but their more rapidly improving multimodal frontier predicts an efficiency crossover within practical pretraining budgets.
These findings position encoder-free architectures as a promising direction, and we expect this work to encourage broader exploration of encoder-free multimodal pretraining.

\section{Preliminaries}
\label{sec:prelim}

\subsection{Estimating Scaling Laws}
\label{sec:prelim:setup}

\noindent\textbf{Problem Definition.}
Let $M$ denote FLOPs per token~\citep{bi2024deepseek} and $D$ the number of objective tokens, so the training budget is $C=MD$.
At a target budget $C$, the compute-optimal allocation~\citep{kaplan2020scaling,hoffmann2022training} for objective $\mathcal{L}$ is the feasible point on $C=MD$ that minimizes loss:
\begin{equation}
M_{\mathrm{opt}}(C), D_{\mathrm{opt}}(C)
= \argmin_{M,\,D} \mathcal{L}(M, D)
\quad \mathrm{s.t.} \quad MD = C.
\end{equation}
Across budgets, these optima follow the compute-optimal allocation law:
\begin{equation}
M_{\mathrm{opt}}(C) \propto C^{a}, \qquad
D_{\mathrm{opt}}(C) \propto C^{b}, \qquad
a + b = 1.
\end{equation}
The corresponding compute-optimal frontiers follow:
\begin{equation}
\label{eq:compute-law}
\mathcal{L}^{*}(C) = E + K\,C^{-\gamma},
\end{equation}
where $\gamma>0$ is the loss--compute exponent, $K>0$ is a fitted prefactor, and $E$ is the entropy floor induced by the data distribution~\citep{hoffmann2022training}.

\noindent\textbf{IsoFLOP Profiles.}
Following Chinchilla~\citep{hoffmann2022training}, an IsoFLOP profile at budget $C$ is obtained by varying $M$, setting $D=C/M$, and fitting validation loss as a quadratic function of $\log M$.
The fitted minimum defines $M_{\mathrm{opt}}(C)$.
The corresponding $D_{\mathrm{opt}}(C)=C/M_{\mathrm{opt}}(C)$ and $\mathcal{L}^{*}(C)$ then follow directly.
Repeating this procedure across budgets provides the optima used to estimate the allocation law and the compute-optimal frontiers.

\subsection{Efficiency Gain}
\label{sec:prelim:eg}

Following MAI-Thinking-1~\citep{microsoftai2026mai}, let $\lambda$ be a loss reachable by both systems and let $C_s(\lambda)$ denote the actual training compute required by system $s \in \{\mathrm{tar},\mathrm{ref}\}$ to reach it under the training regime being compared.
The compute efficiency gain is
\begin{equation}
\label{eq:eg}
\mathrm{EG}^{C}_{\mathrm{tar}\leftarrow\mathrm{ref}}(\lambda) =
C_{\mathrm{ref}}(\lambda) / C_{\mathrm{tar}}(\lambda).
\end{equation}
Further, let $M_s(\lambda)$ denote the FLOPs per token of the model actually used by system $s$ at this point of equal loss.
The model efficiency gain is
\begin{equation}
\label{eq:egm}
\mathrm{EG}^{M}_{\mathrm{tar}\leftarrow\mathrm{ref}}(\lambda) =
M_{\mathrm{ref}}(\lambda) / M_{\mathrm{tar}}(\lambda).
\end{equation}
For both metrics, values above $1.0$ indicate that the target is more efficient than the reference: the target requires less training compute for $\mathrm{EG}^{C}$ or fewer FLOPs per token for $\mathrm{EG}^{M}$ to reach $\lambda$.

\subsection{Model Ladder}
\label{sec:training}
\begin{wrapfigure}{r}{0.50\textwidth}
\centering
\vspace{-10pt}
\includegraphics[width=\linewidth]{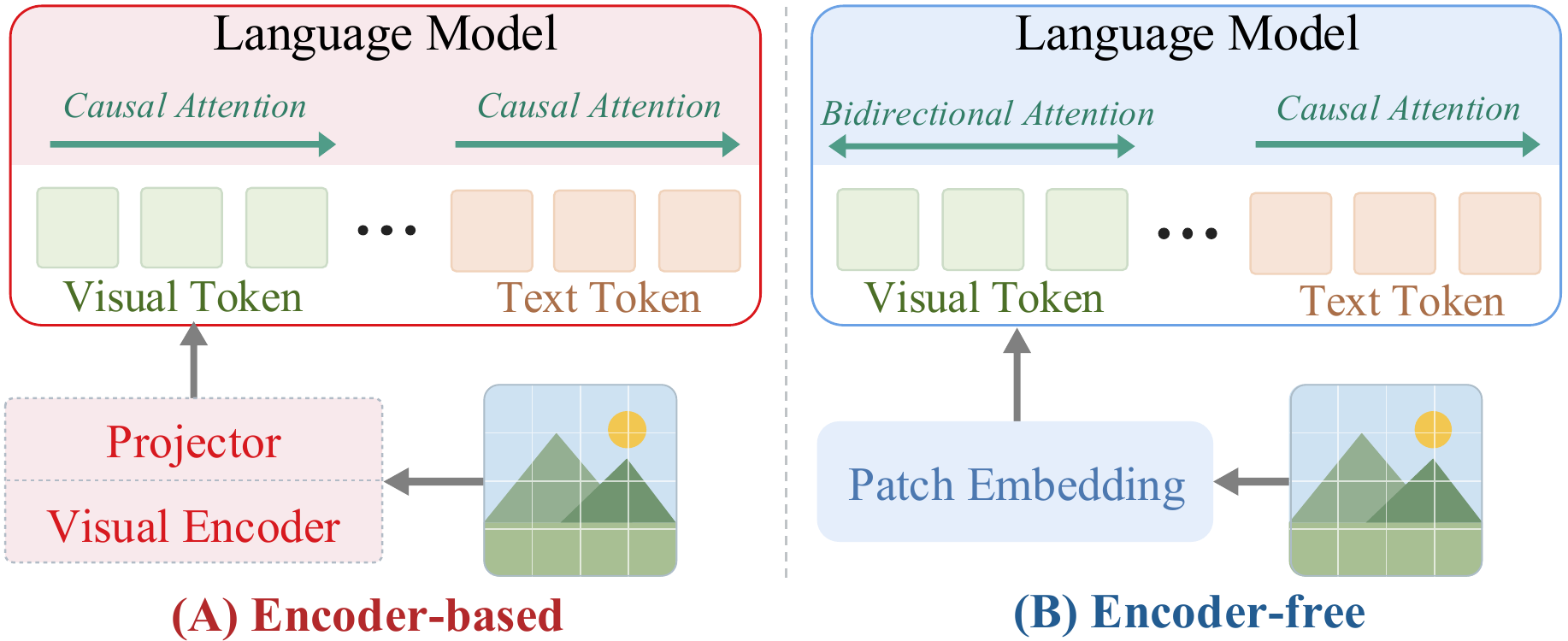}
\vspace{-20pt}
\caption{Architectural comparison of encoder-based and encoder-free MLLMs.}
\label{fig:arch}
\vspace{-5pt}
\end{wrapfigure}
We compare encoder-free and encoder-based MLLMs on a matched ladder of 11 sparse MoE language models with 1.1B--44B total and 71M--2.4B active non-embedding parameters, sharing the same data mixture, optimization setup, and visual-token granularity.
As shown in Fig.~\ref{fig:arch}, the encoder-based model encodes images with a pretrained SigLIP 2 ViT~\citep{tschannen2025siglip2}, followed by a ConvPool adapter and a projector, and applies causal attention to all tokens.
Following common practice~\citep{bai2025qwen3,team2025gemma}, the ViT keeps the same size across decoder scales and is trained jointly with the decoder.
The encoder-free model instead maps raw image patches into the decoder through a patch projection~\citep{team2026gemma}, where visual tokens attend bidirectionally within each image~\citep{team2026gemma, diao2026pixels, tuna2, fang2026let} and all other attention remains causal. We also study a fully causal variant.
More implementation details are provided in Appendix~\ref{app:implementation}.

\subsection{Compute Accounting}
\label{sec:accounting}

The decoder FLOPs per token can be decomposed into a term from matrix multiplications applied to each token, which is independent of sequence length, and a term from self-attention, which grows with sequence length:
\begin{equation}
M_o^{(s)}
=
M_{\mathrm{base}}
+
M_{\mathrm{attn}}^{(s)}\,\ell_o,
\qquad
C_o^{(s)}
=
M_o^{(s)}D_o,
\end{equation}
where $s\in\{\mathrm{free},\mathrm{based}\}$ indexes the model family and $o\in\{\mathrm{text},\mathrm{mm}\}$ indexes the objective. Here, $D_o$ counts all tokens processed in batches for objective $o$ ($D_{\mathrm{mm}}$ includes both visual and text tokens). The mean packed length $\ell_o$ depends on the objective, while $M_{\mathrm{attn}}^{(s)}$ differs between the two families because they use different attention patterns over visual tokens. The fixed expert activation ratio of $8/256$ keeps $M$ approximately proportional to the number of active parameters.

Since we vary decoder scale with a fixed visual encoder, we use decoder FLOPs as the primary compute measure, focusing the analysis on the tradeoff between decoder capacity and training tokens. Appendix~\ref{app:accounting} shows that including visual encoder FLOPs leaves the main conclusions unchanged and further strengthens the relative efficiency of encoder-free models.

\begin{figure}[t]
\centering
\includegraphics[width=\textwidth]{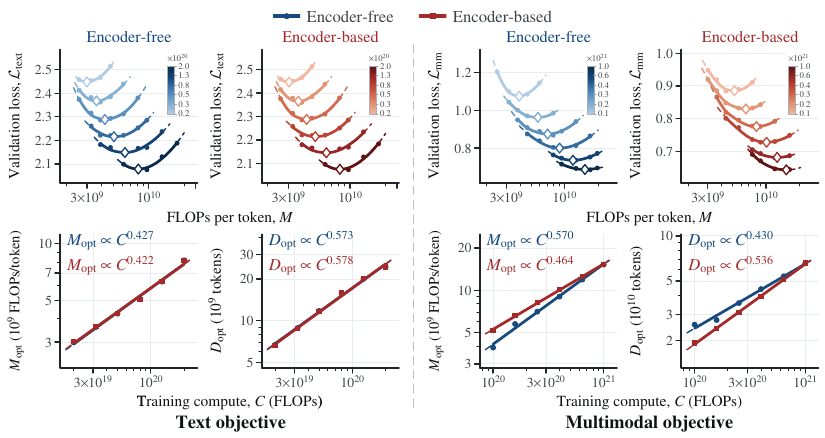}
\vspace{-10pt}
\caption{Estimating compute-optimal allocation with IsoFLOP profiles for the text objective (left) and the multimodal objective (right).}
\label{fig:isoflop-allocation}
\end{figure}

\section{Scaling Laws for Encoder-Free Multimodal Pretraining}
\label{sec:scaling-results}

In this section, we explore how removing the pretrained visual encoder changes the scaling behavior of the language model.
\S\ref{sec:results:allocation} first examines how encoder removal changes compute-optimal allocation.
\S\ref{sec:results:equal-loss} then uses the resulting allocation laws to test whether encoder-free models can catch up at scale, under both compute-optimal allocation and overtraining.
Finally, \S\ref{sec:decoder-consequences} examines how the decoder takes over visual encoding through vision-specific adaptation.

\subsection{How Does Encoder Removal Change Compute-Optimal Allocation?}
\label{sec:results:allocation}

We estimate compute-optimal allocation for both encoder-free and encoder-based models using the IsoFLOP profiles described in \S\ref{sec:prelim:setup}, with the results shown in Fig.~\ref{fig:isoflop-allocation}.
At each budget for each model, the best point in the profile gives $M_{\mathrm{opt}}$ and $D_{\mathrm{opt}}$, whose scaling with $C$ yields the allocation exponents $a$ and $b$.

\begin{wraptable}{r}{0.40\textwidth}
  \centering
  \vspace{-15pt}
  \caption{Model allocation exponent $a$ of encoder-free models under bidirectional and causal attention over visual tokens.}
  \label{tab:allocation-exponents}
  \small
  \setlength{\tabcolsep}{6pt}
  \begin{tabular}{@{}lcc@{}}
  \toprule
  Attention & Text & Multimodal \\
  \midrule
  Bidirectional & $0.427$ & $0.570$ \\
  Causal & $0.436$ & $0.557$ \\
  \bottomrule
  \end{tabular}
  \vspace{-3pt}
\end{wraptable}
\textbf{On the text objective}, the two architectures have nearly identical model allocation exponents ($a=0.427$ for encoder-free and $a=0.422$ for encoder-based models), which is expected.
\textbf{On the multimodal objective}, however, removing the encoder increases $a$ from $0.464$ to $0.570$, indicating that compute-optimal training allocates more compute to model scale.
For these multimodal estimates, a bootstrap gives central $80\%$ intervals of $[0.546,0.595]$ and $[0.458,0.472]$ for the encoder-free and encoder-based models, respectively (Appendix~\ref{app:crossover-bootstrap}).
The shift persists under causal attention over visual tokens, which gives $a=0.557$ (Tab.~\ref{tab:allocation-exponents}; Appendix~\ref{app:causal-allocation}), suggesting that it is not an artifact of the attention mask.
Together, these results suggest that removing the visual encoder increases the decoder's representational burden and favors larger models.

\takeaway{Removing the visual encoder leaves the compute-optimal allocation on the text objective unchanged, but shifts multimodal training toward larger models (Fig.~\ref{fig:isoflop-allocation}).}

\vspace{-2pt}
\subsection{Can Encoder-Free Models Catch Up at Scale?}
\label{sec:results:equal-loss}

After estimating compute-optimal allocation, we explore how much compute and decoder scale encoder-free models need to match encoder-based models at equal validation loss.
For all metrics in this subsection, encoder-free is the target and encoder-based is the reference.
We report two ratios.
$\mathrm{EG}^{C}$ compares the training compute required at equal loss.
Because the ratio is reference over target, values below $1.0$ mean encoder-free models need more compute.
$\mathrm{EG}^{M}$ compares the FLOPs per token selected at the point of equal loss.
Values below $1.0$ mean the encoder-free model at equal loss uses a larger decoder.
We first evaluate both ratios on the compute-optimal frontier, then examine how they shift under overtraining.

\begin{figure}[t]
\centering
\includegraphics[width=0.97\textwidth]{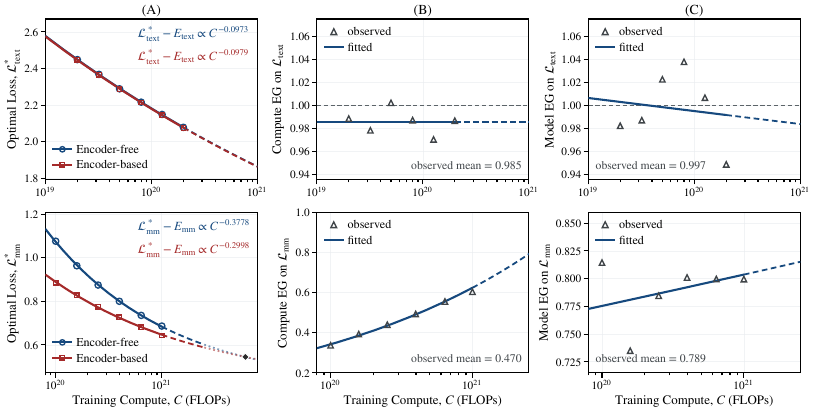}
\vspace{-10pt}
\caption{Efficiency gain analysis on the compute-optimal frontier for the text (top) and multimodal (bottom) objectives.
\textbf{(A)} fitted loss--compute scaling laws, \textbf{(B)} compute efficiency gain ($\mathrm{EG}^{C}$), \textbf{(C)} model efficiency gain ($\mathrm{EG}^{M}$).
}
\label{fig:equal-loss}
\end{figure}

\vspace{-2pt}
\subsubsection{Efficiency Gain Under Compute-Optimal Allocation}
We first compare the two systems on their respective compute-optimal frontiers, where each training budget is allocated between model scale and tokens to minimize loss. Fixing a target loss then determines both the required compute and the corresponding optimal FLOPs per token.
Fig.~\ref{fig:equal-loss} summarizes the fitted loss--compute scaling laws and the resulting efficiency gains.

\noindent\textbf{Text Objective.}
On $\mathcal{L}_{\mathrm{text}}$, the two fitted loss curves are almost indistinguishable.
Their loss--compute exponents are also nearly identical ($0.0973$ versus $0.0979$).
The fitted and measured $\mathrm{EG}^{C}$ values stay around $0.98$, and the fitted $\mathrm{EG}^{C}$ curve is nearly flat.
$\mathrm{EG}^{M}$ at equal loss also stays close to $1.0$, around $0.99$, with only a slight downward drift across the fitted range.
Both deviations remain small and nearly constant across the fitted range, so text acts as a nearly matched control rather than a regime with a meaningful encoder-free penalty.

\noindent\textbf{Multimodal Objective.}
On $\mathcal{L}_{\mathrm{mm}}$, encoder-free models require more training compute to attain the same loss throughout the measured range.
However, this gap narrows with scale: the fitted $\mathrm{EG}^{C}$ increases because the encoder-free loss decreases more rapidly with training compute.
At equal loss, encoder-free models also favor a larger decoder, with $\mathrm{EG}^{M}\approx0.80$, corresponding to approximately $1.25\times$ the FLOPs per token.
Extrapolating the fitted scaling laws places the efficiency crossover on the order of $10^{22}$ FLOPs (Fig.~\ref{fig:equal-loss}A, bottom). The point estimate is $6.1\times10^{21}$ FLOPs, with a conditional bootstrap $80\%$ interval of $[4.2\times10^{21},1.0\times10^{22}]$ (Appendix~\ref{app:crossover-bootstrap}).
This projection assumes that the fitted laws persist beyond the measured range, with the visual encoder held at a fixed size and the irreducible loss determined only by the data distribution.
Even accounting for this uncertainty, the crossover remains roughly three orders of magnitude below the pretraining compute of recent flagship models, e.g., approximately $10^{25}$ FLOPs for Kimi K2.5~\citep{team2026kimi25}.

\subsubsection{Efficiency Gain Under Overtraining}
\label{sec:results:overtraining}
The compute-optimal frontier is not the only practical regime, since deployment models are often overtrained by spending extra tokens at a fixed model scale to reduce inference cost at a target quality~\citep{sardana2023beyond}.
We therefore test whether overtraining changes the efficiency gain.
Starting from the compute-optimal point at base budget $C_{\mathrm{base}}$, overtraining keeps the model scale fixed and trains on $k$ times as many tokens, where $k$ is the overtraining factor, so the actual training compute is $C_{\mathrm{actual}}=kC_{\mathrm{base}}$.

\begin{figure}[t]
 \centering
 \includegraphics[width=0.97\textwidth]{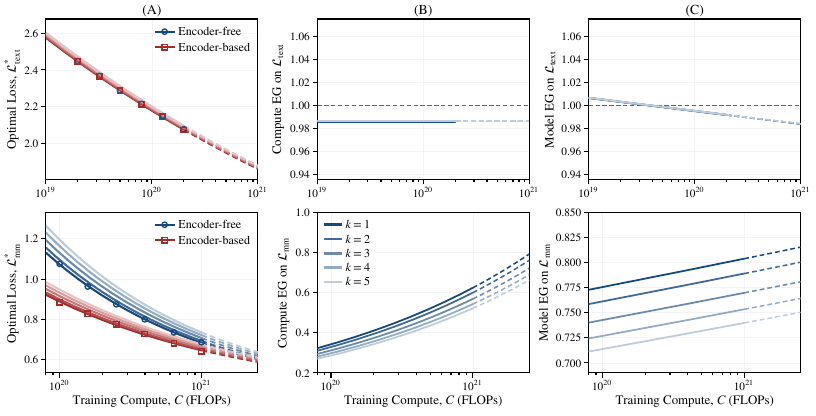}
 \vspace{-10pt}
 \caption{Efficiency gain analysis under overtraining for the text (top) and multimodal (bottom) objectives, with shades denoting overtraining factors $k=1$--$5$ from darkest to lightest.
 \textbf{(A)} loss--compute scaling laws, \textbf{(B)} compute efficiency gain ($\mathrm{EG}^{C}$), \textbf{(C)} model efficiency gain ($\mathrm{EG}^{M}$).
 }
 \label{fig:overtraining}
\end{figure}

We model overtraining as a shift in the prefactor of the loss--compute law, following prior scaling analyses~\citep{gadre2025language}.
Under the separable form $\mathcal{L}(M,D)=E+AM^{-\alpha}+BD^{-\beta}$ with $C_{\mathrm{base}}=MD$, the compute-optimal frontier is $\mathcal{L}^{*}(C_{\mathrm{base}})=E+KC_{\mathrm{base}}^{-\gamma}$ with $\gamma=\alpha\beta/(\alpha+\beta)$.
Fixing $M=M_{\mathrm{opt}}(C_{\mathrm{base}})$ and setting $D=kD_{\mathrm{opt}}(C_{\mathrm{base}})$ gives
\begin{equation}
\label{eq:overtrain}
\mathcal{L}(C_{\mathrm{base}},k)=E+g(k)\,K\,C_{\mathrm{base}}^{-\gamma},
\end{equation}
where the multiplier $g(k)$ depends on $k$ but not on $C_{\mathrm{base}}$.
Overtraining therefore leaves $E$ and $\gamma$ unchanged and rescales only the reducible term (derivation in Appendix~\ref{app:overtraining-proof}).
In practice, we reuse $E$, $K$, and $\gamma$ from the compute-optimal fit and estimate only an empirical $g(k)$ for each $k$ (Appendix~\ref{app:training}), which requires far fewer runs and lets us run overtraining experiments at smaller model scales.

Fig.~\ref{fig:overtraining} shows how overtraining changes the comparison at equal loss.
On the multimodal objective, the shift is asymmetric: the same overtraining factor $k$ changes the two systems' prefactors by different amounts because their data exponents and the optimal split between loss terms differ.
At $k=5$, the extrapolated crossover remains on the order of $10^{22}$ FLOPs but arrives later than under compute-optimal allocation. The point estimate is $1.2\times10^{22}$ FLOPs, with a conditional bootstrap $80\%$ interval of $[8.4\times10^{21},2.0\times10^{22}]$ (Appendix~\ref{app:crossover-bootstrap}). At the largest fitted budget, overtraining also lowers $\mathrm{EG}^{C}$ from $0.62$ to $0.52$ and $\mathrm{EG}^{M}$ from $0.80$ to $0.74$.
This is consistent with the allocation shift in \S\ref{sec:results:allocation}: since encoder-free models favor larger decoders on multimodal data, spending extra compute on tokens at a fixed model scale benefits them less.
By contrast, encoder-free and encoder-based models remain nearly matched on the text objective under overtraining, with both ratios changing by less than $1\%$ from their $k=1$ values.

\takeaway{Both architectures nearly overlap on text loss frontiers, while encoder-free models initially lag on multimodal loss but are predicted to catch up within practical budgets (Figs.~\ref{fig:equal-loss}--\ref{fig:overtraining}).}

\subsubsection{Analysis by Topic}
\label{sec:topic-level-analysis}

\begin{figure}[t]
\centering
\includegraphics[width=\textwidth]{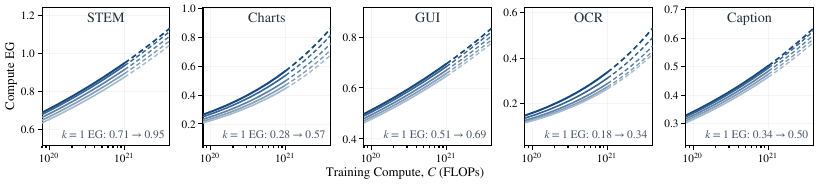}
\vspace{-15pt}
\caption{Compute efficiency gain on each multimodal topic. Curves from dark to light correspond to overtraining factors $k=1$--$5$. Solid segments span the fitted budgets, and dashed segments extrapolate the fitted laws.}
\label{fig:topic-catchup}
\end{figure}

The aggregate multimodal loss summarizes the overall trend but obscures variation across topics.
We therefore repeat the analysis at equal loss on each major multimodal topic (Fig.~\ref{fig:topic-catchup}). Appendix~\ref{app:topic-isoflop} reports the underlying IsoFLOP profiles.

Within our compute range, encoder-free models remain less compute-efficient than encoder-based models on every topic, yet topics differ markedly in the size of the remaining gap and how quickly it narrows with scale.
Extrapolating the fitted laws, encoder-free models catch up first on STEM, which is already close to parity at the largest fitted budget, then on Charts, whose gap narrows quickly, and considerably later on GUI, OCR, and Caption.
This ordering is consistent with how strongly each topic relies on pretrained visual representations.
The STEM subset consists mainly of text, symbols, and simple diagrams, and chart inputs can often be reduced to symbolic content. Once this content is extracted, prediction depends mainly on language and reasoning.
By contrast, captioning requires rich representations of natural images, while GUI and OCR demand detailed spatial and textual perception, for which a pretrained encoder provides a strong prior that encoder-free models must learn from scratch.

\takeaway{The crossover varies by topic, arriving earlier on topics that rely mainly on language (e.g., STEM) and much later on perception-intensive ones (e.g., Caption) (Fig.~\ref{fig:topic-catchup}).}

\subsection{How Does the Decoder Take Over Visual Encoding?}
\label{sec:decoder-consequences}

\begin{figure}[t]
\centering
\includegraphics[width=\linewidth]{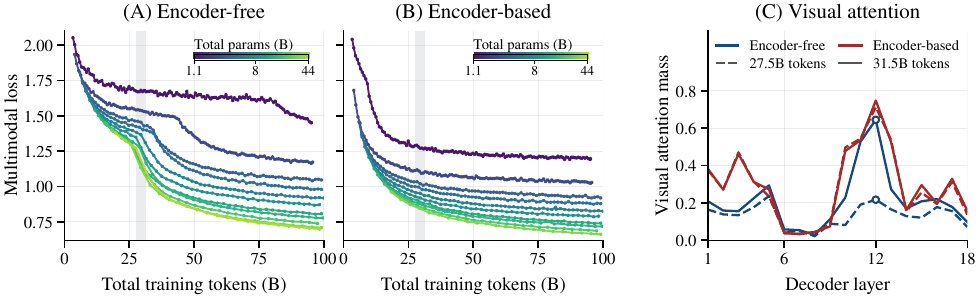}
\vspace{-15pt}
\caption{Visual learning across training tokens. \textbf{(A, B)} Multimodal validation loss during training. \textbf{(C)} Attention mass on visual tokens per decoder layer for the 8B models, before and after the encoder-free loss drop (shaded in A, B).}
\label{fig:mm-transition}
\end{figure}

Removing the visual encoder shifts visual representation learning into the decoder. To trace this shift, we first compare how the multimodal loss of encoder-free and encoder-based models evolves over training tokens, and then examine three probes inside the decoder: attention over visual tokens, layerwise evolution of visual representations, and expert routing.

\noindent\textbf{Emergence of Visual Encoding During Training.}
The multimodal loss of encoder-based models decreases smoothly, whereas that of encoder-free models decreases slowly at first, then drops sharply within a short span (Fig.~\ref{fig:mm-transition}A).
To understand this drop, we compare attention to visual tokens in the 8B models before and after it: at layer 12, the encoder-free model's attention to visual tokens rises from $0.217$ to $0.645$, approaching that of the encoder-based model (Fig.~\ref{fig:mm-transition}C).
We hypothesize that the slow early phase reflects the decoder bootstrapping its own visual representations. Since the loss is applied only to text tokens, visual tokens receive learning signal only when text attends to them. Initially uninformative and thus ignored, they learn slowly until they become useful enough to attract attention, after which learning accelerates. A pretrained encoder supplies useful visual representations from the start and thus avoids this stage, consistent with the high attention to visual tokens in the encoder-based model at both checkpoints.
At every multimodal IsoFLOP budget, the compute-optimal models have already 
passed this drop, so the fitted frontier reflects the smooth regime after it.

\Needspace{18\baselineskip}
\begin{wrapfigure}{r}{0.49\textwidth}
\centering
\vspace{-10pt}
\includegraphics[width=\linewidth]{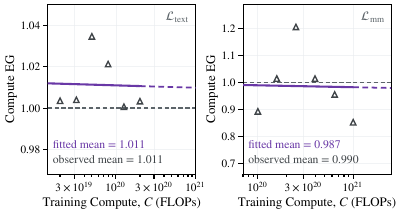}
\vspace{-22pt}
\caption{Comparison between causal and bidirectional attention on visual tokens. Values above $1.0$ favor causal.}
\label{fig:attention-ablation}
\end{wrapfigure}
\noindent\textbf{Encoder-Like Contextualization: Bidirectional Attention.}
We compare bidirectional and causal attention over visual tokens by rerunning the encoder-free ladder with the causal variant (Fig.~\ref{fig:attention-ablation}).
Causal attention is slightly better on the text objective at the measured budgets, reaching the same loss for about $1\%$ less compute ($\mathrm{EG}^{C}=1.011$), but this gain shrinks with compute.
On the multimodal objective, causal attention is mildly worse on average ($\mathrm{EG}^{C}=0.990$), and the multimodal gain of bidirectional attention becomes larger with compute.
This pattern is consistent with a transfer of function from encoder to decoder.
In encoder-based models, the ViT bidirectionally contextualizes image patches before they reach the decoder.
Once the ViT is removed, bidirectional attention among visual tokens allows the decoder to assume part of this role.
The increasing multimodal benefit of bidirectional attention with scale suggests that larger decoders exploit these interactions better, while its diminishing cost on text indicates limited interference with language modeling.

\Needspace{18\baselineskip}
\begin{wrapfigure}{r}{0.49\textwidth}
\centering
\vspace{-10pt}
\includegraphics[width=\linewidth]{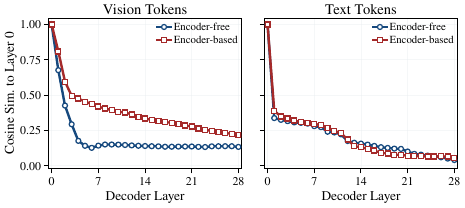}
\vspace{-20pt}
\caption{Layerwise representation evolution. Cosine similarity to the input representation at layer 0 across decoder layers for visual tokens (left) and text tokens (right).}
\vspace{-5pt}
\label{fig:token-layer-similarity}
\end{wrapfigure}
\noindent\textbf{Encoder-Like Early Processing: Layerwise Representation Evolution.}
In encoder-based MLLMs, the ViT has already transformed visual tokens into semantic representations, so they undergo little additional processing in shallow decoder layers~\citep{fan2026visual}.
If the decoder takes over this transformation, its shallow layers should instead rewrite visual tokens substantially.
Fig.~\ref{fig:token-layer-similarity} tests this using cosine similarity between each layer's token representations and their layer-0 inputs.
Without the visual encoder, visual tokens move away from their inputs much earlier (left), while text token trajectories remain close across the two systems (right). The same pattern holds across model scales (Appendix~\ref{app:layer-similarity-scales}).
The shallow decoder layers thus act as an implicit visual encoding stage, performing the transformation that the ViT performs in encoder-based models, and this change is specific to visual tokens.

\begin{figure}[t]
 \centering
 \includegraphics[width=\linewidth]{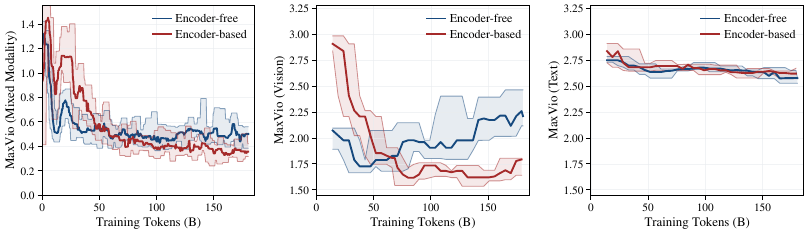}
 \vspace{-10pt}
 \caption{Expert load imbalance over training. MaxVio is computed jointly over visual and text tokens (left), over visual tokens only (middle), and over text tokens only (right). Higher values indicate greater imbalance.}
 \label{fig:load-balancing}
\end{figure}

\noindent\textbf{Encoder-Like Dedicated Capacity: Expert Routing.} Both systems use the same sparse decoder, so routing differences show how the decoder absorbs the changed visual representations.
We quantify expert load imbalance using MaxVio~\citep{wang2024auxiliary}, the relative excess of the most-loaded expert's load over the perfectly balanced load.
As shown in Fig.~\ref{fig:load-balancing}, both systems have similarly low overall MaxVio when visual and text tokens are aggregated, although the encoder-free values are slightly higher.
Separating tokens by modality reveals substantially greater expert load imbalance for both visual and text tokens than the aggregate suggests.
For text tokens, the two architectures remain closely matched.
In contrast, across the four largest model sizes, encoder-free models exhibit consistently higher average MaxVio and a wider band for visual tokens throughout training.
The close match on text argues against a routing shift across the whole model and localizes the effect to visual processing.
This concentration is consistent with the decoder allocating a subset of its experts to play the role of the vision-specific parameters that the ViT previously provided.

\takeaway{The decoder adapts to take over visual encoding: bidirectional attention among visual tokens, shallow-layer visual processing, and concentrated expert routing (Figs.~\ref{fig:attention-ablation}--\ref{fig:load-balancing}).}

\section{Related Work}

\noindent\textbf{Encoder-Free MLLMs.} Encoder-free MLLMs remove the visual encoder and pass projected image patches directly into the decoder.
Early systems such as Fuyu, EVE, and SOLO established the feasibility of this design~\citep{fuyu-8b,diao2024unveiling,chen2024solo}.
Moving beyond feasibility, SAIL systematically studies model and data scalability, cross-modal information flow, and visual representation learning within a single Transformer~\citep{lei2025scalability}.
Later work improves visual competence through alignment or distillation~\citep{tao2025hovle,wang2025vision,li2026breen,yang2025haplovl}, additional capacity dedicated to each modality~\citep{luo2025mono,luo2025mono15,diao2025evev2}, native vision-language primitives~\citep{diao2026pixels}, and extensions to video~\citep{yi2025video,li2025breaking,diao2026pixelsov}, 3D~\citep{tang2026exploring}, segmentation~\citep{zhang2025pixel}, and unified understanding and generation~\citep{team2024chameleon,li2025synergen,li2026onecat,diao2026sensenova,tuna2,wang2026multimodal}.
Recent large systems such as the Gemma~4 12B Unified model~\citep{team2026gemma} and Inkling~\citep{thinkingmachines2026inkling} also explore native multimodal input without a visual encoder.
Our work complements these efforts by comparing behavior between encoder-free and encoder-based MLLMs.

\noindent\textbf{Scaling Laws.} Predicting training behavior at large scale from small runs is now standard practice, spanning loss trends, data scaling, compute-optimal allocation, and training hyperparameters~\citep{kaplan2020scaling,hoffmann2022training,bi2024deepseek,bjorck2025scaling,li2025predictable,clark2022unified,alabdulmohsin2022revisiting,wang2026smooth}.
Recent work extends this methodology to native multimodal pretraining.
In particular, the compute-optimal data requirement grows faster for vision than for language in unified multimodal pretraining~\citep{tong2026beyond}, and native MLLMs trained under data constraints exhibit coupled scaling between the visual encoder and the language model~\citep{tian2026navil}.
Others compare early and late fusion trained from scratch~\citep{shukor2025scaling} or study how the data mixture affects the scaling of encoder-free MoE models~\citep{wu2026scaling}.
In contrast, we compare encoder-free models against an encoder-based baseline on a shared MoE decoder ladder, predict the compute budget at which encoder-free models catch up under both compute-optimal allocation and overtraining, and analyze how the decoder takes over visual encoding via vision-specific adaptation.

\section{Conclusion}

Encoder-free MLLMs are less compute-efficient on the multimodal objective at the scales we evaluate. However, their compute-optimal loss decreases faster as compute increases. Our fitted scaling laws predict a crossover on the order of $10^{22}$ FLOPs under compute-optimal allocation and at a higher budget under $5\times$ overtraining. Encoder-free scaling also favors larger decoders, delays the decoder's reliance on visual input, strengthens interactions among visual tokens, shifts visual representation transformation earlier, and concentrates expert routing. In contrast, text scaling remains largely unchanged. These findings point to future work on decoder architectures and training strategies that improve the compute efficiency of native visual input.

\section*{Acknowledgments}

We would like to thank Yan Fang for many fruitful and insightful discussions throughout the course of this work.

\bibliography{references}
\bibliographystyle{plainnat}

\clearpage
\appendix
\raggedbottom

\section*{Appendix}

\etocdepthtag.toc{appendix}
\etocsettagdepth{mainmatter}{none}
\etocsettagdepth{appendix}{subsection}
\etocsettocstyle{}{}
\begingroup
\hypersetup{linkcolor=takeawayaccent}
\tableofcontents
\endgroup
\clearpage

\section{Implementation Details}
\label{app:implementation}

\subsection{Visual Front End Architectures}
\label{app:visual-front-ends}

\noindent\textbf{Encoder-Free Front End.} Both front ends produce one visual token per $32\times32$ pixel region. Our encoder-free front end adapts the patch projection pipeline used in the Gemma~4 12B Unified model~\citep{team2026gemma}. In the original pipeline, $16\times16$ patches are merged in $3\times3$ spatial groups, yielding visual tokens that each cover a $48\times48$ pixel region. To ensure a controlled comparison, we instead use $2\times2$ merging, matching the encoder-based front end in both visual-token granularity and token count. Each merged patch of raw pixels passes through a LayerNorm, a linear layer, and a second LayerNorm. Learned factorized 2D position embeddings, obtained by summing separate embeddings for the two spatial axes, are then added, followed by a third LayerNorm, an RMSNorm, and a linear projection to the decoder width.

\noindent\textbf{Encoder-Based Front End.} The encoder-based front end uses a pretrained SigLIP 2 ViT~\citep{tschannen2025siglip2} with 27 layers, width 1152, patch size 16, and AnyRes processing. A $2\times2$ ConvPool adapter and a projector match the encoder-free visual-token granularity. We choose this roughly 400M encoder scale as a representative practical setting used by recent advanced MLLMs~\citep{team2025gemma,bai2025qwen3,team2026kimi25}. The ViT is trained jointly with the decoder in every run, while its architecture and parameter count remain fixed across decoder scales.

\noindent\textbf{Matched Comparison Protocol.} Both front ends use the same preprocessing pipeline, consume the same pixels, and pass the same number of visual tokens to the decoder. The main comparison therefore holds visual content and token count fixed while varying the representation attached to each token and the attention pattern over visual tokens described in \S\ref{sec:decoder-consequences}. Fixing the ViT across the ladder isolates decoder scaling and avoids introducing joint scaling of the encoder and decoder as an additional variable. Consequently, the ViT accounts for a smaller fraction of total model capacity as the decoder grows, and the reported scaling trends are conditional on this regime with a fixed encoder size. Jointly scaling the visual encoder would define a different allocation problem, requiring a separate sweep that balances potential representation gains against additional compute in the front end.

\subsection{Decoder Architecture and Model Ladder}
\label{app:model-ladder}

The two systems share a complete ladder of 11 rungs. Model depth and width vary across rungs, while the MoE topology remains fixed. Every rung uses $256$ routed experts, activates the top $8$ experts for each token, and includes one shared expert~\citep{dai2024deepseekmoe}. The activation ratio of routed experts is therefore fixed at $8/256=1/32$ across model scales, keeping all rungs within a consistent architectural family. The first layer is dense, followed by MoE layers.
All models use RMSNorm~\citep{zhang2019root} and RoPE~\citep{su2024roformer}.

\subsection{Training Setup}
\label{app:training}

\noindent\textbf{Optimization and Hyperparameters.}
All runs use a sequence length of $4{,}096$, the Muon optimizer~\citep{jordan2024muon}, and $2{,}000$ warmup steps. Training hyperparameters are selected as a function of model scale using an internal scaling law. In particular, the batch size and learning rate vary across the model ladder. At each rung, identical hyperparameters are used for the encoder-based and encoder-free MLLMs.

\noindent\textbf{Overtraining Runs and Multiplier Fitting.}
For both model families, we train compute-optimal model sizes at $k\in\{2,3,4,5\}$. To allow for deviations from the idealized separable loss model underlying Eq.~\ref{eq:overtrain}, we keep $E$, $K_s$, and $\gamma_s$ fixed to their compute-optimal estimates and fit an empirical multiplier $g_s^{\mathrm{emp}}(k)$ using the available overtraining runs. At $k=5$, the fitted multipliers are $0.645$ for encoder-free and $0.692$ for encoder-based models. These empirical estimates are used for the overtraining analysis in Fig.~\ref{fig:overtraining}.

\subsection{Training and Validation Data}
\label{app:data}

\noindent\textbf{Training Mixture.} The corpus is a $1{:}1$ mixture of text and multimodal data. Since we focus on architectural scaling rather than effects of the data mixture, we keep this ratio fixed rather than treating it as a study variable. The balanced mixture provides sufficient multimodal training tokens for robust estimation of multimodal scaling. Multimodal sources cover tasks such as captioning, charts, grounding, GUI, OCR, STEM, and knowledge. Text sources cover domains such as STEM, code, books, and wikis.

\noindent\textbf{Validation Losses.} Validation data are disjoint from training and identical across systems. We separately track losses on pure text and on multimodal topics: text losses are computed on sequences containing no visual tokens, whereas multimodal losses are computed over text prediction tokens in sequences conditioned on images, with visual tokens masked out from the loss, and are grouped by topic. These masked visual positions remain included in $D_{\mathrm{mm}}$ and in decoder FLOP accounting. The primary validation view weights major topics uniformly while preserving the training mixture weights of sources within each topic.

\subsection{Scaling Law Fitting Setup}
\label{app:fitting-setup}

Following prior scaling law work~\citep{kaplan2020scaling,hoffmann2022training}, we use held-out validation loss as the primary metric and fit separate laws for the text and multimodal objectives, $\mathcal{L}_{\mathrm{text}}$ and $\mathcal{L}_{\mathrm{mm}}$.
IsoFLOP fits use six logarithmically spaced budgets for each objective, covering $2 \times 10^{19}$ to $2 \times 10^{20}$ FLOPs for $\mathcal{L}_{\mathrm{text}}$ and $1 \times 10^{20}$ to $1 \times 10^{21}$ FLOPs for $\mathcal{L}_{\mathrm{mm}}$.

\section{Robustness of the Scaling Law Estimates}
\label{app:robustness}

In this section, we assess the robustness of the scaling law estimates.
We begin by checking the IsoFLOP estimator against a validation curve envelope estimator in Appendix~\ref{app:envelope}.
Next, we evaluate held-out extrapolation accuracy in Appendix~\ref{app:forecasting}.
Appendix~\ref{app:E-sensitivity} tests the sensitivity of the loss--compute fits to the shared irreducible loss.
Appendix~\ref{app:crossover-bootstrap} quantifies conditional fitting uncertainty with a residual bootstrap that preserves pairs matched by budget.
\subsection{Comparison with the Validation Curve Envelope}
\label{app:envelope}

This subsection compares the IsoFLOP estimates with a validation curve envelope.
We first describe the estimator, then report the text and multimodal fits in Tab.~\ref{tab:estimator-exponents}.

\noindent\textbf{Validation Curve Envelope.}
We adapt the envelope approach of Chinchilla~\citep{hoffmann2022training}, which uses training loss curves, to validation loss trajectories: the held-out validation loss of each run, measured at its intermediate checkpoints. At each of 200 logarithmically spaced compute budgets, we linearly interpolate each trajectory in log compute and fit a local quadratic in $\log M$ to the five model scales around the lowest observed loss on the ladder. The quadratic vertex gives a continuous estimate of the optimum derived from the trajectories. Because adjacent budgets are highly correlated, we aggregate these estimates by their median within 24 equal bins in log compute before fitting the allocation and loss laws. We use Huber regression in log space for the allocation laws to limit the influence of isolated errors from local interpolation. Unlike the six IsoFLOP profiles, this estimator constructs a dense frontier from points along the validation trajectories.

\begin{table}[t]
\centering
\caption{Scaling exponents estimated by IsoFLOP and validation curve envelope fitting.}
\label{tab:estimator-exponents}
\small
\setlength{\tabcolsep}{5pt}
\begin{tabular}{llcccc}
\toprule
Objective & System & Estimator & Loss $\gamma$ & Model $a$ & Data $b$ \\
\midrule
Text & Encoder-free  & IsoFLOP  & 0.0973 & 0.427 & 0.573 \\
Text & Encoder-free  & Envelope & 0.0905 & 0.430 & 0.570 \\
Text & Encoder-based & IsoFLOP  & 0.0979 & 0.422 & 0.578 \\
Text & Encoder-based & Envelope & 0.0915 & 0.435 & 0.565 \\
\midrule
Multimodal & Encoder-free  & IsoFLOP  & 0.3778 & 0.570 & 0.430 \\
Multimodal & Encoder-free  & Envelope & 0.3668 & 0.583 & 0.417 \\
Multimodal & Encoder-based & IsoFLOP  & 0.2998 & 0.464 & 0.536 \\
Multimodal & Encoder-based & Envelope & 0.3050 & 0.475 & 0.525 \\
\bottomrule
\end{tabular}
\end{table}

\begin{figure}[t]
\centering
\includegraphics[width=\linewidth]{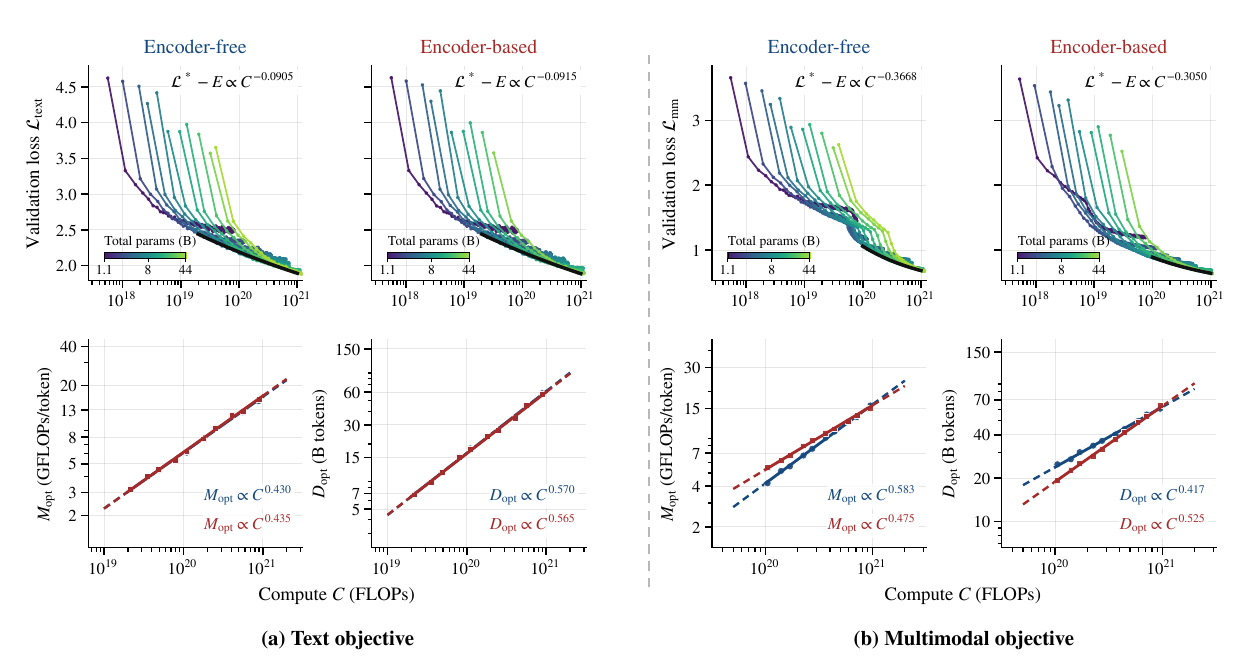}
\vspace{-20pt}
\caption{Validation curve envelopes grouped by objective: (a) text and (b) multimodal. Within each group, the top row shows unsmoothed encoder-free and encoder-based validation trajectories. The bottom row shows representative estimates of compute-optimal $M_{\mathrm{opt}}$ and $D_{\mathrm{opt}}$ derived from the trajectories and binned in log compute, with fitted allocation laws.}
\label{fig:envelope}
\end{figure}

\noindent\textbf{Text Objective.}
Fig.~\ref{fig:envelope}(a) shows the text fits. Over its broader trajectory range, the envelope estimator gives $a=0.430$ for encoder-free and $a=0.435$ for encoder-based models, close to the IsoFLOP values of $0.427$ and $0.422$. Its loss--compute exponents of $0.0905$ and $0.0915$ are also close to the IsoFLOP estimates.

\noindent\textbf{Multimodal Objective.}
Fig.~\ref{fig:envelope}(b) shows the corresponding multimodal fits. The envelope estimator gives $a=0.583$ for encoder-free and $a=0.475$ for encoder-based models, close to the IsoFLOP values of $0.570$ and $0.464$. Its loss--compute exponents of $0.3668$ and $0.3050$ are also close to the IsoFLOP values of $0.3778$ and $0.2998$, preserving the same ordering: the encoder-free frontier is steeper than the encoder-based frontier.

\subsection{Extrapolation Error Analysis}
\label{app:forecasting}

We test whether the compute-optimal fits forecast held-out IsoFLOP optima when the target budget is a modest multiple of the largest fitted budget.
For each objective, we fit separate loss--compute scaling laws for each architecture to the first four IsoFLOP optima, then evaluate the forecast at a held-out budget beyond the fitting window.
Held-out optima are estimated independently with the quadratic IsoFLOP procedure from \S\ref{sec:prelim:setup}.
All parameters of each architecture are estimated using only the fitting subset at lower compute.

\begin{figure}[t]
  \centering
  \includegraphics[width=0.80\linewidth]{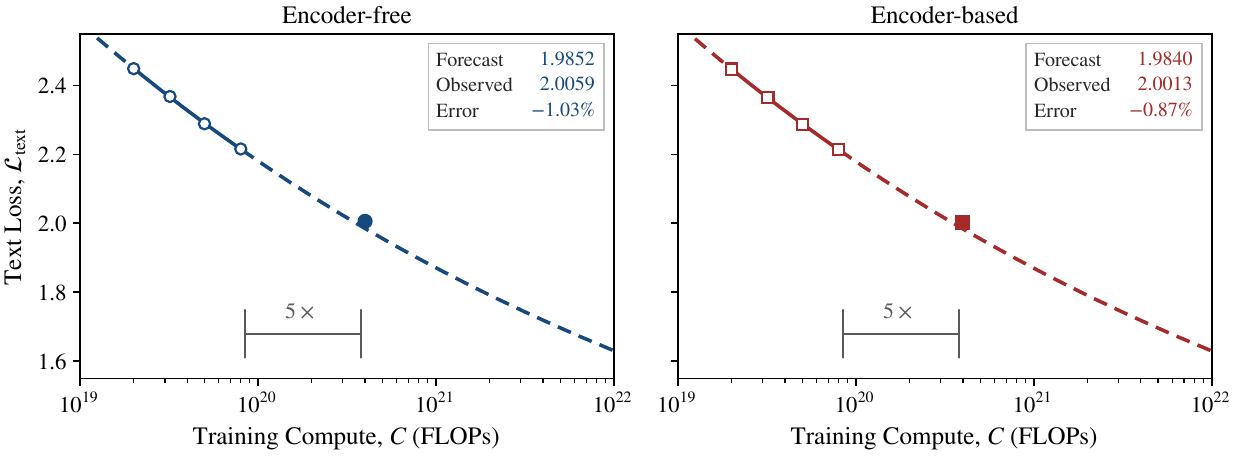}
  \vspace{-10pt}
  \caption{Extrapolation error analysis on text loss.
  Laws fitted through $8\times10^{19}$ FLOPs are evaluated against held-out IsoFLOP optima at $4\times10^{20}$ FLOPs, a $5\times$ extrapolation.
  Dashed segments indicate extrapolation.}
  \label{fig:extrapolation-validation}
  \vspace{-10pt}
\end{figure}

\noindent\textbf{Text Objective.}
For this check, we additionally run a separate IsoFLOP profile at $4\times10^{20}$ FLOPs, outside the main fitting range of the scaling laws, and use it only as a held-out extrapolation target.
The forecasting fit at lower compute spans $2\times10^{19}$ to $8\times10^{19}$ FLOPs, making the held-out target a $5\times$ extrapolation beyond the largest fitted budget.
The held-out profiles contain seven encoder-free and six encoder-based model scales, with both estimated minima lying inside the sampled model range.
Fig.~\ref{fig:extrapolation-validation} shows signed relative errors of $-1.03\%$ and $-0.87\%$.

\begin{figure}[t]
  \centering
  \includegraphics[width=0.80\linewidth]{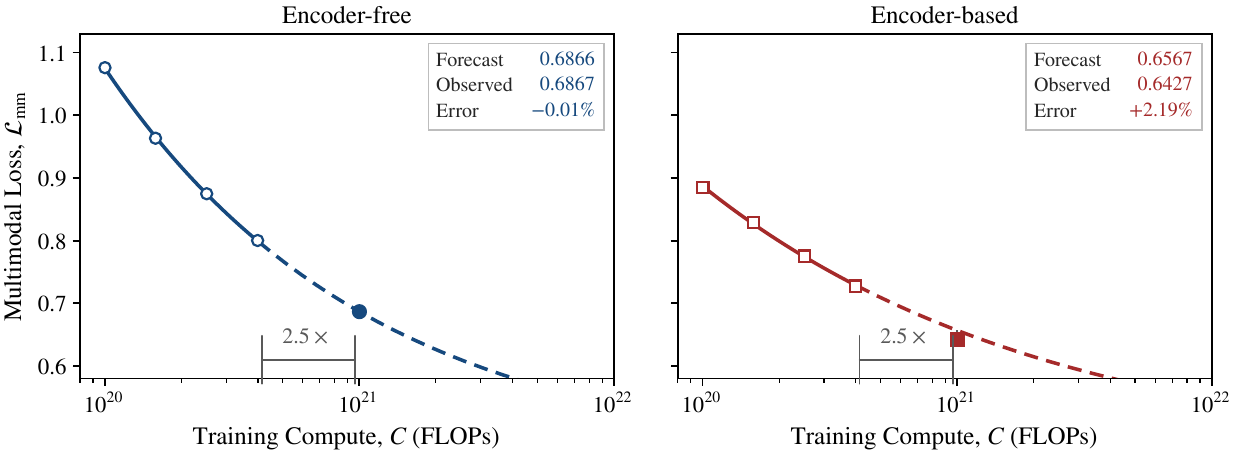}
  \vspace{-10pt}
  \caption{Extrapolation error analysis on multimodal loss.
  Laws fitted through $4\times10^{20}$ FLOPs are evaluated against held-out IsoFLOP optima at $1\times10^{21}$ FLOPs, a $2.5\times$ extrapolation.
  Dashed segments indicate extrapolation.}
  \label{fig:extrapolation-validation-mm}
\end{figure}

\noindent\textbf{Multimodal Objective.}
The fitting window spans IsoFLOP optima through $4\times10^{20}$ FLOPs.
We forecast the optimum at $1\times10^{21}$ FLOPs, a $2.5\times$ extrapolation beyond the largest fitted budget.
Fig.~\ref{fig:extrapolation-validation-mm} shows signed relative errors of $-0.01\%$ for encoder-free models and $+2.19\%$ for encoder-based models. Across both objectives, the same fitting procedure forecasts held-out compute-optimal losses over these extrapolation factors with relative errors within about $2\%$.

\subsection{Sensitivity to the Irreducible Loss}
\label{app:E-sensitivity}

\begin{figure}[t]
  \centering
  \includegraphics[width=0.96\linewidth]{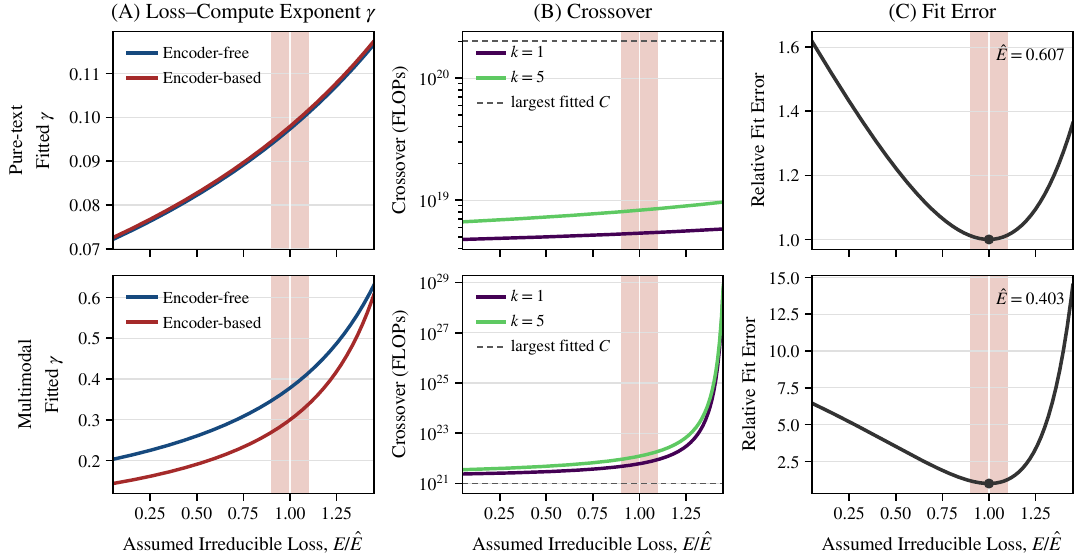}
  \vspace{-10pt}
  \caption{Sensitivity to the irreducible loss of each objective. Each row fixes $E_o/\hat E_o$ and jointly refits the encoder-free and encoder-based loss--compute scaling laws with a common $E_o$. \textbf{(A)} Fitted loss--compute exponent $\gamma$. \textbf{(B)} Implied crossover compute under compute-optimal allocation ($k=1$) and $5\times$ overtraining ($k=5$). The dashed line marks the largest fitted budget. \textbf{(C)} Joint SSE on raw loss, normalized by its minimum. The red band marks $\pm10\%$ around the fitted floor.}
  \label{fig:E-sensitivity}
\end{figure}

This subsection specifies the fit with a common $E$ used for the loss--compute laws, then tests how the estimates move when $E$ is perturbed (Fig.~\ref{fig:E-sensitivity} and Tab.~\ref{tab:E-sensitivity}).

\noindent\textbf{Common Irreducible Loss.}
Following prior scaling law work~\citep{henighan2020scaling,hoffmann2022training}, we interpret the irreducible term $E$ as the conditional entropy floor induced by the data distribution and prediction objective. Because the encoder-free and encoder-based systems are trained and evaluated on the same examples with the same next-token objective, their Bayes-optimal loss floor is shared. We therefore impose one $E_o$ per objective across architectures, while allowing the finite compute terms $K_s$ and $\gamma_s$ to differ.

This constraint of a common $E$ is a structural assumption, not a claim that our observations at finite scale identify the asymptote. Its interpretation further assumes that both model families can eliminate approximation error that depends on architecture as scale grows. In our setting, with six multimodal frontier points per system over approximately one decade of compute, $E$ trades off strongly with the fitted slope. Separate floors for each architecture can therefore absorb differences within the finite range without reliably identifying distinct entropy limits. More broadly, asymptotic parameters of scaling laws are known to be sensitive to the fitting sample and specification~\citep{besiroglu2024chinchilla,choshen2024hitchhiker,porian2024resolving}. We consequently treat the constraint as theoretically motivated and test below which conclusions depend on it.

For objective $o$ and system $s\in\{\mathrm{free},\mathrm{based}\}$, we jointly solve
\begin{equation}
\min_{E_o,\{K_s,\gamma_s\}}
\sum_s\sum_i
\left[
\mathcal{L}_{s,i}
-E_o-K_s C_{s,i}^{-\gamma_s}
\right]^2,
\label{eq:shared-E-fit}
\end{equation}
where $K_s>0$ and $\gamma_s>0$. The observations $(C_{s,i},\mathcal{L}_{s,i})$ are the compute-optimal IsoFLOP vertices. The fitted floors are $\hat E_{\mathrm{text}}=0.607$ and $\hat E_{\mathrm{mm}}=0.403$.

\noindent\textbf{Sensitivity.}
We then fix $E_o$ on a grid from $0.05\hat E_o$ to $1.45\hat E_o$ and, at every grid point, refit $(K_s,\gamma_s)$ by constrained least squares. The two objectives are swept independently. Fig.~\ref{fig:E-sensitivity} reports the resulting loss--compute exponents, crossover points, and joint fitting error.
The exponent ordering does not depend on the assumed floor: on the multimodal objective, $\gamma_{\mathrm{free}}$ exceeds $\gamma_{\mathrm{based}}$ at every grid point, whereas the two text exponents remain nearly identical throughout. Within $\pm10\%$ of the fitted floor, $E_{\mathrm{mm}}/\hat E_{\mathrm{mm}}\in[0.90,1.10]$, the multimodal exponent gap stays between $0.077$ and $0.078$ while the joint fitting error rises by at most $27\%$.
The gap between the two objectives is likewise not an artifact of an overestimated floor. Lowering $E$ reduces every fitted exponent, yet even at $E_o=0.05\hat E_o$, close to a pure power law, the encoder-based multimodal exponent ($0.144$) remains twice the text exponent ($0.073$).
The fitted exponent values themselves, and any extrapolated crossover, are more sensitive to $E$ (Tab.~\ref{tab:E-sensitivity}). These are sensitivity ranges rather than statistical confidence intervals.

\begin{table}[H]
\centering
\caption{Multimodal sensitivity to perturbations of the fitted shared irreducible loss.}
\label{tab:E-sensitivity}
\small
\setlength{\tabcolsep}{6pt}
\renewcommand{\arraystretch}{1.08}
\begin{tabular}{lcccc}
\toprule
$E_{\mathrm{mm}}/\hat E_{\mathrm{mm}}$ &
$E_{\mathrm{mm}}$ &
$\gamma_{\mathrm{free}}-\gamma_{\mathrm{based}}$ &
Compute-optimal &
$5\times$ overtraining \\
\midrule
$[0.90,1.10]$ &
$[0.363,0.444]$ &
$[0.077,0.078]$ &
$[4.8,9.0]\times10^{21}$ &
$[0.9,1.9]\times10^{22}$ \\
\bottomrule
\end{tabular}
\end{table}

\subsection{Conditional Bootstrap Uncertainty}
\label{app:crossover-bootstrap}

In this section, we use a conditional residual bootstrap with paired resampling across architectures to characterize fitting uncertainty in the multimodal loss--compute and allocation exponents, their differences between architectures, and the extrapolated crossover budgets under compute-optimal allocation and $5\times$ overtraining.

\noindent\textbf{Bootstrap Protocol.}
For each architecture, we compute the residuals between the six estimated optimal losses from the IsoFLOP profiles and the corresponding values of the fitted compute law, then center them by subtracting the mean for each architecture.
At each bootstrap replicate, we sample six matched residual pairs with replacement and add them to the fitted losses at the six fixed compute budgets.
We then jointly refit the two loss--compute scaling laws by least squares on the untransformed loss scale, estimating a common $E_{\mathrm{mm}}$ anew in each replicate.
This paired resampling preserves the association at each budget between the encoder-free and encoder-based residuals.

We apply the same sampled budget indices to the centered residual pairs from the fits of the allocation laws.
Each bootstrap replicate then yields estimates of the loss--compute and allocation exponents, their differences between architectures, and the crossover budgets under compute-optimal allocation and $5\times$ overtraining.
For the $5\times$ crossover, each replicate reuses the original estimates of the multipliers $g_s(5)$ rather than estimating them again.

Following the reporting convention of Chinchilla~\citep{hoffmann2022training}, we report central $80\%$ bootstrap percentile intervals, bounded by the $10$th and $90$th percentiles of the bootstrap distribution.

\noindent\textbf{Loss--Compute Exponents.}
On the multimodal objective, the encoder-free loss--compute exponent has a bootstrap median of $0.3781$, with a central $80\%$ interval of $[0.3686,0.3873]$, whereas the encoder-based exponent has a median of $0.3002$, with an interval of $[0.2882,0.3112]$.
Their difference, $\Delta\gamma=\gamma_{\mathrm{free}}-\gamma_{\mathrm{based}}$, has a median of $0.0781$ and an interval of $[0.0687,0.0876]$.
The difference is positive across all $4{,}000$ conditional bootstrap replicates.

\begin{figure}[t]
  \centering
  \includegraphics[width=\textwidth]{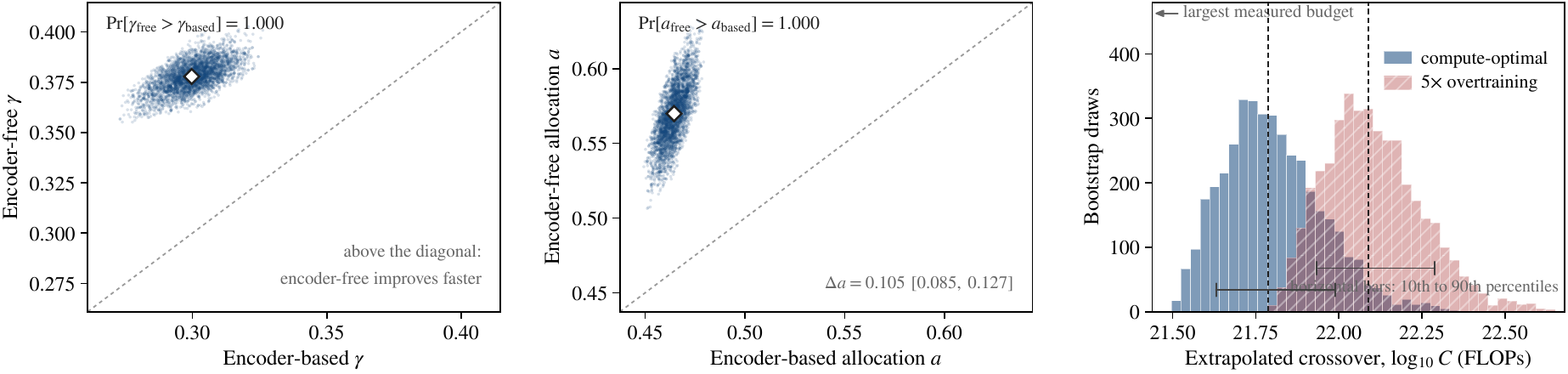}
  \vspace{-15pt}
  \caption{Conditional bootstrap uncertainty of the multimodal scaling fits and extrapolated crossover.
  \textbf{Left:} joint distribution of the encoder-free and encoder-based loss--compute exponents $\gamma$.
  \textbf{Middle:} joint distribution of their model allocation exponents $a$.
  \textbf{Right:} distributions of the extrapolated crossover compute under compute-optimal allocation and $5\times$ overtraining.
  Horizontal bars span the $10$th to $90$th percentiles, and dashed lines mark point estimates.
  }
  \label{fig:crossover-bootstrap}
\end{figure}

\noindent\textbf{Compute-Optimal Allocation Exponents.}
The model allocation exponent $a_{\mathrm{free}}$ has a bootstrap median of $0.570$, with a central $80\%$ interval of $[0.546,0.595]$, whereas $a_{\mathrm{based}}$ has a median of $0.465$, with an interval of $[0.458,0.472]$.
Their difference, $\Delta a=a_{\mathrm{free}}-a_{\mathrm{based}}$, has a median of $0.105$ and an interval of $[0.085,0.127]$, and is positive across all conditional bootstrap replicates.
The corresponding data exponents have medians of $0.430$ and $0.536$ for the encoder-free and encoder-based systems, with central $80\%$ intervals of $[0.405,0.454]$ and $[0.528,0.543]$, respectively.
The left and middle panels of Fig.~\ref{fig:crossover-bootstrap} show the joint bootstrap distributions, while Tab.~\ref{tab:bootstrap-exponents} summarizes the marginal estimates and differences between architectures.

\begin{table}[t]
\centering
\caption{Conditional bootstrap uncertainty of the multimodal scaling exponents. Estimates are the fitted IsoFLOP exponents. Intervals are central $80\%$ bootstrap percentile intervals ($10$th to $90$th percentiles).}
\label{tab:bootstrap-exponents}
\small
\setlength{\tabcolsep}{7pt}
\renewcommand{\arraystretch}{1.10}
\begin{tabular}{lcccc}
\toprule
& \multicolumn{2}{c}{Encoder-free} & \multicolumn{2}{c}{Encoder-based} \\
\grouprule{2-3}\grouprule{4-5}
Exponent & Estimate & $80\%$ interval & Estimate & $80\%$ interval \\
\midrule
Loss $\gamma$ & $0.3778$ & $[0.3686,\,0.3873]$ & $0.2998$ & $[0.2882,\,0.3112]$ \\
Model $a$ & $0.570$ & $[0.546,\,0.595]$ & $0.464$ & $[0.458,\,0.472]$ \\
Data $b$ & $0.430$ & $[0.405,\,0.454]$ & $0.536$ & $[0.528,\,0.543]$ \\
\bottomrule
\end{tabular}
\end{table}

\noindent\textbf{Crossover.}
The compute-optimal crossover has a median of $6.1\times10^{21}$ FLOPs and an interval of $[4.2\times10^{21},1.0\times10^{22}]$, and the crossover under $5\times$ overtraining has a median of $1.2\times10^{22}$ FLOPs and an interval of $[8.4\times10^{21},2.0\times10^{22}]$.
All replicates produce finite crossovers within $[10^{21},10^{23}]$ FLOPs, and refits that each leave out one budget place the compute-optimal crossover between $4.3\times10^{21}$ and $9.8\times10^{21}$ FLOPs.
The right panel of Fig.~\ref{fig:crossover-bootstrap} shows the two distributions.

\section{Compute Accounting for the Visual Front End}
\label{app:accounting}

This appendix adds the training cost of each visual front end. Let $\phi$ denote the resulting ViT share of encoder-based training compute. Because the loss of every trained configuration is unchanged, we estimate the encoder-based frontier under full accounting from a parametric fit $\mathcal{L}(M,D)=E+AM^{-\alpha}+BD^{-\beta}$ to its IsoFLOP points, minimizing it over $M$ at each total budget.

\subsection{Compute-Optimal Allocation}
\label{app:accounting-allocation}

Under full accounting, the resulting compute-optimal allocation follows $M_{\mathrm{opt}}\propto C^{0.273}$ and $D_{\mathrm{opt}}\propto C^{0.727}$ for encoder-based models, compared with $a=0.464$ and $b=0.536$ when only decoder FLOPs are counted, while the encoder-free exponents remain $a=0.570$ and $b=0.430$ (Tab.~\ref{tab:accounting}). Because $M_{\mathrm{full}}=M_{\mathrm{dec}}+\mathrm{const}$, we have $d\ln M_{\mathrm{full}}/d\ln M_{\mathrm{dec}}=1-\phi$, so the encoder-based exponent measured in total FLOPs per token is compressed while $\phi$ is large and approaches the decoder-accounting value as $\phi\to0$. Under both conventions, encoder-free training still favors larger models.

\subsection{Loss--Compute Exponent}
\label{app:accounting-invariance}

Refitting $\mathcal{L}^{*}=E+KC^{-\gamma}$ to the six encoder-based optima under full accounting, with the shared $\hat E_{\mathrm{mm}}$ fixed, gives $\gamma_{\mathrm{based}}=0.362$ with a root mean square error of $0.003$, while the encoder-free exponent remains $0.378$ (Tab.~\ref{tab:accounting}). The exponent gap narrows from $0.078$ to $0.016$ but keeps its sign. The narrowing is a finite-scale effect: charging the ViT raises the cost of small encoder-based runs relatively more than that of large ones, which steepens the frontier in total compute. As $\phi\to0$ at larger scales, the encoder-based exponent approaches the decoder-accounting value of $0.300$, and the gap approaches $0.078$.

\subsection{Summary}
\label{app:accounting-summary}

Neither conclusion of the main text depends on the accounting convention. For allocation, encoder-free training favors larger models than encoder-based training under both conventions, and encoder-free models retain the larger loss--compute exponent. For the crossover, charging the fixed ${\sim}400$M ViT only adds cost to encoder-based runs and leaves the encoder-free frontier unchanged, so it shifts every comparison at equal loss toward encoder-free models. At the crossover under decoder accounting in \S\ref{sec:results:equal-loss}, encoder-free models therefore already require less total compute, and the crossover under full accounting occurs no later than $6.1\times10^{21}$ FLOPs. We do not refit a numerical crossover under this convention.

\begin{table}[H]
\centering
\caption{Multimodal scaling exponents under decoder and full compute accounting.}
\label{tab:accounting}
\small
\setlength{\tabcolsep}{8pt}
\renewcommand{\arraystretch}{1.12}
\begin{tabular}{llccc}
\toprule
Accounting & System & Model $a$ & Data $b$ & Loss $\gamma$ \\
\midrule
Decoder only & Encoder-free  & $0.570$ & $0.430$ & $0.378$ \\
Decoder only & Encoder-based & $0.464$ & $0.536$ & $0.300$ \\
\midrule
Full & Encoder-free  & $0.570$ & $0.430$ & $0.378$ \\
Full & Encoder-based & $0.273$ & $0.727$ & $0.362$ \\
\bottomrule
\end{tabular}
\end{table}

\section{Derivation of \texorpdfstring{Eq.~\ref*{eq:overtrain}}{the Overtraining Loss Equation}}
\label{app:overtraining-proof}

We derive Eq.~\ref{eq:overtrain} from \S\ref{sec:results:overtraining}.
Here $k$ multiplies the compute-optimal token count at fixed model scale: $M=M_{\mathrm{opt}}(C_{\mathrm{base}})$, $D=kD_{\mathrm{opt}}(C_{\mathrm{base}})$, and $C_{\mathrm{actual}}=kC_{\mathrm{base}}$.

Eliminating $D$ via $D=C_{\mathrm{base}}/M$, the excess loss at fixed $C_{\mathrm{base}}$ becomes
\[
\mathcal{L}(M,C_{\mathrm{base}}/M)-E
=AM^{-\alpha}+BC_{\mathrm{base}}^{-\beta}M^\beta.
\]
This expression is strictly convex in $\log M$, so its unique stationary
point is the global optimum. Differentiating with respect to $M$ and setting the derivative to zero yields
\[
-\alpha A M^{-\alpha-1}+\beta B C_{\mathrm{base}}^{-\beta}M^{\beta-1}=0,
\]
hence
\[
M^{\alpha+\beta}=\frac{\alpha A}{\beta B}\,C_{\mathrm{base}}^{\beta}.
\]
Writing $\rho=\alpha+\beta$ and $m=(\alpha A/\beta B)^{1/\rho}$, we obtain
\[
M_{\mathrm{opt}}(C_{\mathrm{base}})=mC_{\mathrm{base}}^{\beta/\rho}.
\]
The corresponding token count is $D_{\mathrm{opt}}(C_{\mathrm{base}})=C_{\mathrm{base}}/M_{\mathrm{opt}}(C_{\mathrm{base}})$, which simplifies to
\[
D_{\mathrm{opt}}(C_{\mathrm{base}})=m^{-1}C_{\mathrm{base}}^{\alpha/\rho}.
\]

Keeping the model scale fixed at $M_{\mathrm{opt}}(C_{\mathrm{base}})$ and multiplying the optimal token count
by $k$, direct substitution yields
\begin{align*}
\mathcal{L}(C_{\mathrm{base}},k)-E
&=A\!\left(mC_{\mathrm{base}}^{\beta/\rho}\right)^{-\alpha}
{}+B\!\left(km^{-1}C_{\mathrm{base}}^{\alpha/\rho}\right)^{-\beta}\\
&=\left(Am^{-\alpha}+Bm^\beta k^{-\beta}\right)C_{\mathrm{base}}^{-\gamma},
\end{align*}
where $\gamma=\alpha\beta/\rho$. Let
\[
H(k)=Am^{-\alpha}+Bm^\beta k^{-\beta}.
\]
The loss--compute scaling law on the compute-optimal frontier at $k=1$ implies $H(1)=K$. Thus, defining the
dimensionless multiplier $g(k)=H(k)/H(1)$ gives
\[
\mathcal{L}(C_{\mathrm{base}},k)=E+g(k)KC_{\mathrm{base}}^{-\gamma}.
\]
Since $H(k)$ depends on $k$ but not on $C_{\mathrm{base}}$, overtraining changes only the prefactor and leaves the compute exponent unchanged.

\section{Probes Inside the Decoder}
\label{app:decoder-probes}

\subsection{Layerwise Visual Representation Evolution Across Scales}
\label{app:layer-similarity-scales}
\label{app:probe-protocol}

\begin{figure}[H]
  \centering
  \includegraphics[width=\textwidth]{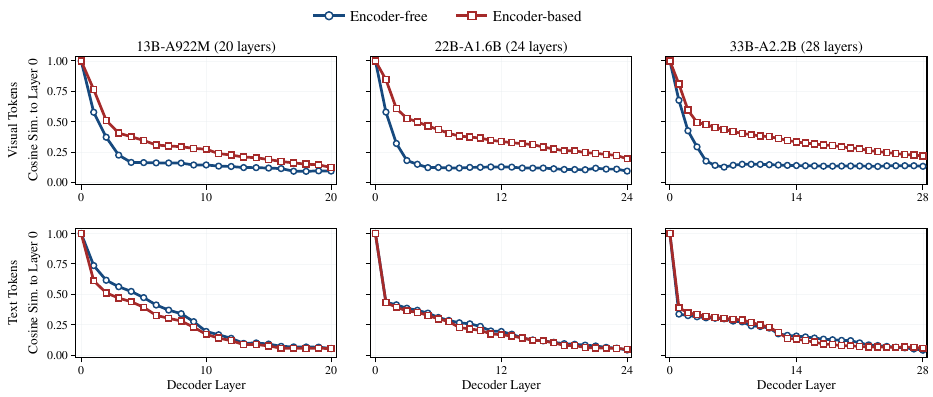}
  \vspace{-20pt}
  \caption{Layerwise similarity to the decoder input across three scales. Encoder-free visual states diverge earlier at every scale, while text trajectories remain closely matched.}
  \label{fig:layer-similarity-scales}
\end{figure}

\S\ref{sec:decoder-consequences} shows that, without a visual encoder, visual tokens move away from their layer-$0$ inputs at shallower decoder layers, while text token trajectories remain close across the two systems. Fig.~\ref{fig:layer-similarity-scales} repeats this comparison at $20$, $24$, and $28$ layers. The same pattern holds at every scale: encoder-free visual states diverge earlier, whereas the text curves stay closely matched. The earlier visual rewriting is therefore not an artifact of a single model size.

\subsection{Compute-Optimal Allocation Under Causal Attention}
\label{app:causal-allocation}

To test whether bidirectional attention among visual tokens causes the encoder-free allocation shift in \S\ref{sec:results:allocation}, we repeat the IsoFLOP procedure of \S\ref{sec:prelim:setup} on the causal encoder-free runs from Fig.~\ref{fig:attention-ablation}. The two settings differ only in how visual tokens attend within each image. The data mixture, optimizer, and training schedule are otherwise matched, and we do not retune hyperparameters for causal attention.

\begin{figure}[H]
  \centering
  \includegraphics[width=0.97\textwidth]{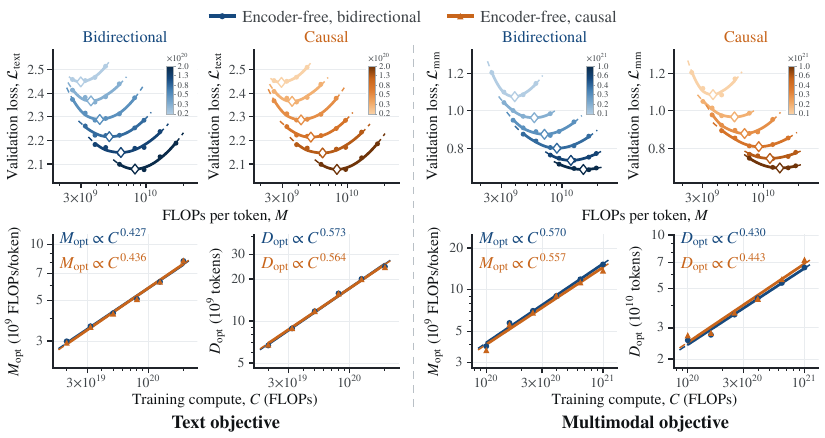}
  \vspace{-10pt}
  \caption{Compute-optimal allocation of encoder-free models under bidirectional and causal attention over visual tokens, for the text objective (left) and the multimodal objective (right). The top row shows IsoFLOP profiles, and the bottom row shows the fitted $M_{\mathrm{opt}}$ and $D_{\mathrm{opt}}$ laws. Diamonds mark fitted vertices and dashed segments indicate extrapolation.}
  \label{fig:causal-isoflop}
\end{figure}

Fig.~\ref{fig:causal-isoflop} compares the two attention settings. Under causal attention, the text objective gives $M_{\mathrm{opt}}\propto C^{0.436}$ and $D_{\mathrm{opt}}\propto C^{0.564}$, while the multimodal objective gives $M_{\mathrm{opt}}\propto C^{0.557}$ and $D_{\mathrm{opt}}\propto C^{0.443}$. These $M$ exponents differ by only $0.009$ and $0.013$ from the bidirectional results ($0.427$ for text and $0.570$ for multimodal). The largest multimodal vertex is extrapolated and should be read with caution. Even so, the multimodal fit still favors much larger models, so the encoder-free allocation trend is not explained by bidirectional attention among visual tokens.

\begin{figure}[H]
  \centering
  \includegraphics[width=0.97\textwidth]{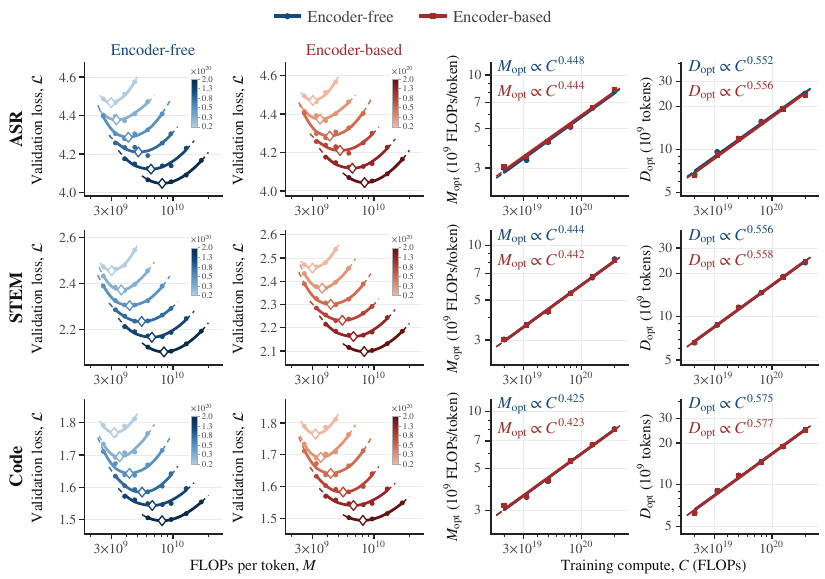}
  \vspace{-10pt}
  \caption{IsoFLOP profiles and fitted compute-optimal allocation laws for three pure text sets. The encoder-free and encoder-based exponents remain close on every topic.}
  \label{fig:topic-isoflop-ood}
\end{figure}

\begin{figure}[H]
  \centering
  \includegraphics[width=0.97\textwidth]{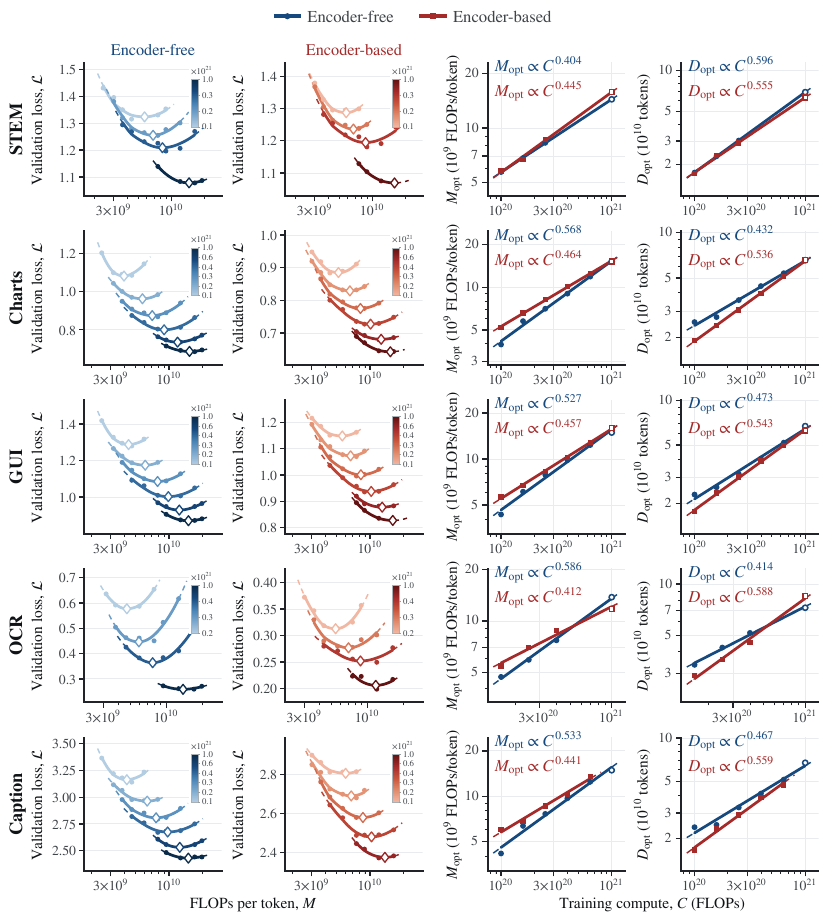}
  \vspace{-10pt}
  \caption{IsoFLOP profiles by topic and fitted compute-optimal allocation laws for five multimodal topics. On most topics, encoder-free models favor larger model scale.}
  \label{fig:topic-isoflop}
\end{figure}

\section{IsoFLOP Profiles by Topic}
\label{app:topic-isoflop}

The main text fits allocation laws to the aggregate validation loss. Here we repeat the analysis on each validation topic separately. No new models are trained. Instead, we evaluate the existing checkpoints on each topic and fit IsoFLOP profiles as in \S\ref{sec:prelim:setup}. Because the model ladder was chosen for the aggregate loss, the optimum for some topics falls near or beyond its edge. We keep mild extrapolations, shown with open markers, and drop vertices that lie far outside the sampled range.

\subsection{Pure Text Topics}
\label{app:topic-isoflop-text}

We start with pure text sets, where the visual encoder plays no role and the two architectures should behave alike. Fig.~\ref{fig:topic-isoflop-ood} confirms this on ASR, STEM, and code. The allocation laws of the two architectures nearly coincide, with $M$ exponents differing by at most $0.004$. This control indicates that the differences in the multimodal topics below come from how visual inputs are processed, not from the two training runs themselves.

\subsection{Multimodal Topics}
\label{app:topic-isoflop-mm}

We now turn to the five multimodal topics from the main text (Fig.~\ref{fig:topic-isoflop}). Unlike the text sets, these topics show a clear gap between the two architectures. On most of them, encoder-free models again favor larger $M$ and smaller $D$, matching the aggregate result, although the size of the shift varies.

\section{Downstream Evaluation}
\label{app:downstream-eval}

Our scaling analyses are based on validation loss. To check whether the same trends hold on downstream tasks, we evaluate the pretrained checkpoints on a set of multimodal benchmarks, without any further training.

\subsection{Benchmarks}
\label{app:downstream-eval-benchmarks}

\paragraph{Perception.}
CV-Bench~\citep{tong2024cambrian} probes basic visual abilities, covering two-dimensional spatial relationships and counting as well as three-dimensional depth and distance. POPE~\citep{li2023evaluating} measures object hallucination by asking whether a queried object appears in the image, and we pool its random, popular, and adversarial subsets. MME~\citep{fu2026mme} poses concise yes/no questions spanning perception and cognition. We report accuracy on all three benchmarks.

\paragraph{Document Understanding.}
ChartQA~\citep{masry2022chartqa} requires extracting values from charts and reasoning over them visually or logically. We report relaxed accuracy, which applies the standard relative tolerance to numerical answers. DocVQA~\citep{mathew2021docvqa} evaluates question answering over document images, where both textual content and layout matter. We report average normalized Levenshtein similarity, which gives partial credit to near-miss answers. AI2D~\citep{kembhavi2016diagram} tests reasoning about the components and relationships in scientific diagrams, and we report multiple-choice accuracy. TextVQA~\citep{singh2019towards} requires reading text in natural images and relating it to the surrounding scene, and we report the VQA consensus score against human answers.

\paragraph{General VQA.}
RealWorldQA~\citep{xai2024realworldqa} tests understanding of real-world scenes, including images captured from vehicles. MMStar~\citep{chen2024we} consists of questions curated so that answering them requires visual evidence, spanning a range of perception and reasoning abilities. MMBench~\citep{liu2024mmbench} offers broad coverage of multimodal capabilities, and we use its English development set. ScienceQA-IMG~\citep{lu2022learn} is the subset of ScienceQA whose school-level science questions come with images. We evaluate answer selection on it without providing the reference explanations. We report accuracy on all four benchmarks.

\subsection{Evaluation Protocol}
\label{app:downstream-eval-protocol}

We evaluate every pretrained checkpoint in the same 3-shot setting: each query is preceded by three solved examples, and all models share the same examples, prompt, and image preprocessing. The examples are drawn from the training split when one exists. Otherwise, we draw them from three images in the evaluation set and exclude all questions about those images from scoring. Images are resized with their aspect ratio preserved, to at most 192 visual tokens per example and 768 for the query, and the full prompt is kept within $4{,}096$ tokens. We use no external OCR, reference explanations, or test-time fine-tuning. For multiple-choice and yes/no questions, the model selects the candidate answer to which it assigns the highest likelihood. For open-ended questions, it decodes greedily for up to 32 tokens.

\subsection{Results}
\label{app:downstream-eval-results}

\begin{figure}[H]
\centering
\includegraphics[width=0.62\textwidth]{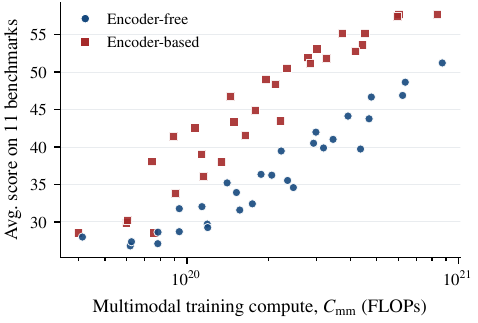}
\vspace{-8pt}
\caption{Downstream benchmark performance against multimodal training compute. The y-axis is the unweighted average 3-shot score over the 11 benchmarks.}
\label{fig:downstream-eval}
\end{figure}

Encoder-free models still score below encoder-based models at the scales we evaluate, but the gap narrows as training compute increases (Fig.~\ref{fig:downstream-eval}). At the largest token budget, the gap also tends to narrow as the model grows (Tab.~\ref{tab:downstream-eval-scores}). Both trends are consistent with the loss-based findings.

\begin{table}[H]
\centering
\caption{3-shot benchmark scores at about 100B training tokens. ``Based'' and ``Free'' denote encoder-based and encoder-free models, respectively.}
\label{tab:downstream-eval-scores}
\small
\setlength{\tabcolsep}{3.2pt}
\begin{tabular}{@{}lcccccccccccc@{}}
\toprule
 & \multicolumn{2}{c}{6.1B-A425M} & \multicolumn{2}{c}{8B-A577M} & \multicolumn{2}{c}{13B-A922M} & \multicolumn{2}{c}{16B-A1B} & \multicolumn{2}{c}{22B-A1.6B} & \multicolumn{2}{c}{33B-A2.2B} \\
\cmidrule(lr){2-3}\cmidrule(lr){4-5}\cmidrule(lr){6-7}\cmidrule(lr){8-9}\cmidrule(lr){10-11}\cmidrule(lr){12-13}
Benchmark & Based & Free & Based & Free & Based & Free & Based & Free & Based & Free & Based & Free \\
\midrule
\rowcolor{customgray}
\multicolumn{13}{c}{\textit{Perception}} \\
CV-Bench & 42.5 & 46.9 & 48.2 & 43.5 & 46.1 & 50.4 & 49.5 & 44.1 & 51.5 & 48.7 & 56.2 & 53.8 \\
POPE & 66.8 & 60.3 & 64.9 & 61.5 & 75.2 & 57.0 & 69.9 & 53.0 & 67.2 & 66.7 & 71.6 & 61.8 \\
MME & 62.1 & 58.0 & 64.3 & 53.1 & 60.9 & 57.1 & 67.7 & 57.9 & 67.9 & 58.7 & 60.7 & 58.8 \\
\rowcolor{customgray}
\multicolumn{13}{c}{\textit{Document}} \\
ChartQA & 34.8 & 10.1 & 48.5 & 29.7 & 51.9 & 29.4 & 47.4 & 35.6 & 53.1 & 37.8 & 54.3 & 42.7 \\
DocVQA & 39.6 & 15.4 & 59.8 & 36.4 & 67.3 & 46.5 & 63.9 & 47.4 & 70.6 & 51.0 & 68.9 & 57.0 \\
AI2D & 44.8 & 40.8 & 46.5 & 41.7 & 50.2 & 45.7 & 50.7 & 47.2 & 55.4 & 49.5 & 57.3 & 53.5 \\
TextVQA & 21.2 & 21.5 & 52.9 & 22.2 & 54.7 & 33.6 & 44.4 & 23.7 & 58.7 & 39.5 & 49.9 & 43.8 \\
\rowcolor{customgray}
\multicolumn{13}{c}{\textit{General VQA}} \\
RealWorldQA & 28.5 & 23.2 & 47.1 & 35.4 & 39.4 & 36.1 & 41.6 & 43.7 & 43.4 & 47.1 & 42.7 & 39.9 \\
MMStar & 32.2 & 27.3 & 34.4 & 28.3 & 35.8 & 28.1 & 36.9 & 29.5 & 35.8 & 29.2 & 37.6 & 31.7 \\
MMBench-EN & 55.6 & 40.1 & 55.5 & 41.6 & 64.7 & 46.3 & 60.8 & 47.2 & 67.1 & 48.3 & 69.1 & 57.6 \\
ScienceQA-IMG & 50.5 & 47.1 & 49.1 & 52.0 & 60.5 & 55.3 & 57.0 & 52.3 & 63.9 & 58.8 & 66.1 & 62.9 \\
\midrule
Average & 43.5 & 35.5 & 51.9 & 40.5 & 55.2 & 44.1 & 53.6 & 43.8 & 57.7 & 48.7 & 57.7 & 51.2 \\
\bottomrule
\end{tabular}
\end{table}

\end{document}